\documentclass[times,twocolumn,final]{elsarticle}  
\DeclareUnicodeCharacter{FF0C}{,}  

\usepackage{medima}
\usepackage{framed,multirow}

\usepackage{amssymb}
\usepackage{latexsym}

\usepackage{url}
\usepackage{xcolor}
\usepackage{adjustbox}

\usepackage{cite}
\usepackage{amsmath,amssymb,amsfonts}
\usepackage{graphicx}
\usepackage{textcomp}

\usepackage{algorithmicx}               
\usepackage{algorithm} 
\usepackage{listings}
\usepackage{algpseudocode}  

\usepackage{array}
\usepackage{booktabs}
\usepackage{colortbl}
\definecolor{dark-gray}{gray}{0.7} 
\usepackage{multirow}

\usepackage{cuted}
\usepackage{graphicx, caption}
\usepackage{color,soul}
\usepackage{rotating}

\usepackage{pdfpages}  
\usepackage{hyperref}  

\definecolor{newcolor}{rgb}{.8,.349,.1}

\journal{Medical Image Analysis}

\renewcommand{\arraystretch}{0.9} 

\begin{document}
\verso{Deng \textit{et~al.}}

\begin{frontmatter}
\title{KPIs 2024 Challenge: Advancing Glomerular Segmentation from Patch- to Slide-Level}%
\vspace{-2.5em}
\author[1,25]{Ruining \snm{Deng}}
\author[1]{Tianyuan \snm{Yao}}
\author[2]{Yucheng \snm{Tang}}
\author[1]{Junlin \snm{Guo}}
\author[1]{Siqi \snm{Lu}}
\author[1]{Juming \snm{Xiong}}
\author[1]{Lining \snm{Yu}}

\author[4]{Quan Huu \snm{Cap}}

\author[5,6,7]{Pengzhou \snm{Cai}}
\author[5]{Libin \snm{Lan}}
\author[8]{Ze \snm{Zhao}}

\author[9]{Adrian \snm{Galdran}}

\author[10]{Amit \snm{Kumar}}
\author[10]{Gunjan \snm{Deotale}}
\author[10]{Dev Kumar \snm{Das}}

\author[11]{Inyoung \snm{Paik}}
\author[11]{Joonho \snm{Lee}}
\author[11]{Geongyu \snm{Lee}}

\author[12]{Yujia \snm{Chen}}
\author[12]{Wangkai \snm{Li}}
\author[12]{Zhaoyang \snm{Li}}

\author[13]{Xuege \snm{Hou}}
\author[13]{Zeyuan \snm{Wu}}
\author[13]{Shengjin \snm{Wang}}

\author[14]{Maximilian \snm{Fischer}}
\author[14]{Lars \snm{Krämer}}

\author[18]{Anghong \snm{Du}}
\author[18]{Le \snm{Zhang}}

\author[9]{Maria Sánchez \snm{Sánchez}}
\author[9]{Helena Sánchez \snm{Ulloa}}
\author[9]{David Ribalta \snm{Heredia}}
\author[9]{Carlos Pérez de Arenaza \snm{García}}

\author[19]{Shuoyu \snm{Xu}}
\author[19]{Bingdou \snm{He}}

\author[20]{Xinping \snm{Cheng}}
\author[20]{Tao \snm{Wang}}

\author[21]{Noémie \snm{Moreau}}
\author[21]{Katarzyna \snm{Bozek}}

\author[23]{Shubham \snm{Innani}}
\author[23]{Ujjwal \snm{Baid}}

\author[24]{Kaura Solomon \snm{Kefas}}

\author[1]{Bennett A. \snm{Landman}}
\author[3]{Yu \snm{Wang}}
\author[3]{Shilin \snm{Zhao}}
\author[3]{Mengmeng \snm{Yin}}
\author[3]{Haichun \snm{Yang}}
\author[1]{Yuankai \snm{Huo}\corref{cor1}}
\cortext[cor1]{Corresponding author. Email: yuankai.huo@vanderbilt.edu}
\address[1]{Vanderbilt University, Nashville, TN 37215, USA}
\address[25]{Weill Cornell Medicine, New York, NY 10021, USA}
\address[2]{NVIDIA Corporation, Redmond, WA 98052, USA}
\address[3]{Vanderbilt University Medical Center, Nashville, TN 37232, USA}
\address[4]{Aillis, Inc., Tokyo 1010042, Japan}
\address[5]{Chongqing University of Technology, Chongqing 400054, China}
\address[6]{Chongqing Zhijian Life Technology Co. LTD, Chongqing 400039, China}
\address[7]{Jinfeng Laboratory，Chongqing 401329, China}
\address[8]{Institute of Computing Technology, Chinese Academy of Sciences, Beijing 100190, China}
\address[9]{Universitat Pompeu Fabra, Ciutat Vella, Barcelona 08002, Spain}
\address[10]{Aira Matrix Private Limited, Thane, Maharashtra 400604, India}
\address[11]{Deep Bio Inc., Research Team, Seoul, KR 08380, Republic of Korea}
\address[12]{University of Science and Technology of China, Hefei, 230026, China}
\address[13]{Tsinghua University \& Beijing National Research Center for Information Science and Technology, Beijing, 100084, China}
\address[14]{German Cancer Research Center,Heidelberg 69120, Germany}
\address[18]{University of Birmingham, Birmingham, B15 2TT, UK}
\address[19]{Bio-totem Pte Ltd, Suzhou 215000, China}
\address[20]{Nanjing University of Science and Technology, Nanjing, Jiangsu 210094, China}
\address[21]{University of Cologne, Cologne 50931, Germany}
\address[23]{Indiana University, Indianapolis, IN 46202, USA}
\address[24]{Xi’an Jiaotong University, Xi’an, Shaanxi 710049, China}

\received{xxxxxx}
\finalform{xxxxxx}
\accepted{xxxxxx}
\availableonline{xxxxxx}
\communicated{xxxxxx}

\begin{abstract}
Chronic kidney disease (CKD) is a major global health issue, affecting over 10\% of the population and causing significant mortality. While kidney biopsy remains the gold standard for CKD diagnosis and treatment, the lack of comprehensive benchmarks for kidney pathology segmentation hinders progress in the field. To address this, we organized the Kidney Pathology Image Segmentation (KPIs) Challenge, introducing a dataset that incorporates preclinical rodent models of CKD with over 10,000 annotated glomeruli from 60+ Periodic Acid Schiff (PAS)-stained whole slide images. The challenge includes two tasks, patch-level segmentation and whole slide image segmentation and detection, evaluated using the Dice Similarity Coefficient (DSC) and F1-score. By encouraging innovative segmentation methods that adapt to diverse CKD models and tissue conditions, the KPIs Challenge aims to advance kidney pathology analysis, establish new benchmarks, and enable precise, large-scale quantification for disease research and diagnosis.
\vspace{-1em}
\end{abstract}

\end{frontmatter}

\section{Introduction}

Chronic kidney disease (CKD) represents a significant global health challenge, causing more deaths annually than breast and prostate cancer combined~\citep{cirillo2024global,iglesias2012gonadal}. Affecting over 10\% of the global population, CKD impacts approximately 800 million individuals worldwide. Kidney biopsy, through both open and percutaneous methods, remains the gold standard for diagnosing and guiding the treatment of CKD~\citep{hogan2016native,agarwal2013basics}. In pathological image analysis, especially within the context of kidney disease, tissue segmentation plays a pivotal role in enabling large-scale, computer-assisted quantification in pathology~\citep{bengtsson2017computer, marti2021digital, gomes2021building}. This analysis provides clinical value for disease diagnosis~\citep{mounier2002cortical}, severity assessment~\citep{kellum2008acute}, and treatment effectiveness evaluation~\citep{jimenez2006mast}.

The advent of deep learning has revolutionized kidney pathology image segmentation~\citep{kumar2017dataset, ding2020multi, ren2017computer, bel2018structure, zeng2020identification}, yet significant challenges remain. Chief among these is the lack of comprehensive benchmarks for developing and evaluating segmentation methods. Existing public datasets for kidney pathology segmentation are limited, primarily containing samples from healthy patients~\footnote{\url{https://www.kaggle.com/competitions/hubmap-hacking-the-human-vasculature/data}}~\footnote{\url{https://www.neptune-study.org}}. This limitation arises because tissue samples are typically obtained through needle biopsies, yielding only small samples. Consequently, there is a pressing need to establish extensive kidney pathology datasets encompassing diverse CKD models.

To address these limitations, we organized the Kidney Pathology Image Segmentation (KPIs) Challenge, expanding the dataset by incorporating preclinical animal models, specifically whole kidney sections from diseased rodents. This competition aims to identify state-of-the-art segmentation methods for glomerular identification across various CKD models. Participants are tasked with developing algorithms capable of accurately segmenting glomeruli at the pixel level, adapting to different CKD models and tissue conditions. This includes distinguishing glomeruli from surrounding tissue components under diverse preparation scenarios, showcasing both the versatility and precision of their approaches. Furthermore, the challenge encourages innovative solutions to address potential obstacles such as variations in glomeruli size, shape, and structural integrity resulting from disease states or preparation techniques.

\begin{figure*}[ht]
\centering

\includegraphics[width=0.6\linewidth]{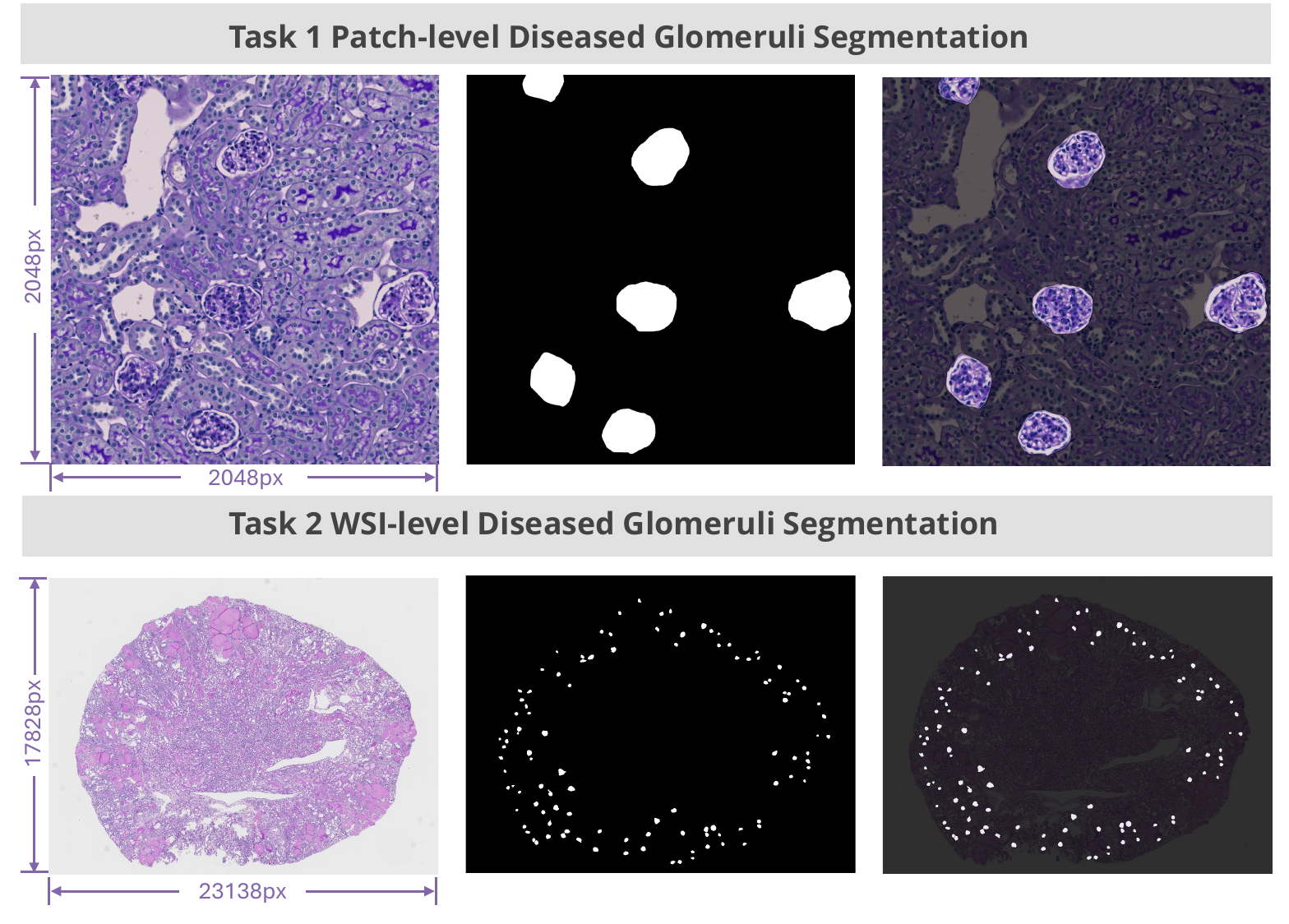}

\caption{\textbf{Overview of the challenge task setup.} The KPIs challenge involves two tasks: (1) Patch-level Diseased Glomeruli Segmentation, focusing on the precise delineation of glomeruli in high-resolution image patches, and (2) Whole Slide Image (WSI)-level Diseased Glomeruli Segmentation, requiring robust segmentation across entire tissue sections. Example images illustrate the input data, ground truth masks, and overlay visualizations for each task. The dataset includes diverse mouse models to highlight morphological and pathological variations.}
\label{fig: challenge}
\end{figure*}

To comprehensively evaluate participants' models, the challenge features two distinct tasks (as shown in Fig.~\ref{fig: challenge}) aimed at advancing the field of kidney pathology image analysis:
\begin{enumerate}
    \item \textbf{Patch-Level Segmentation:} Segmentation of glomeruli within specific image patches.
    \item \textbf{Whole Slide Image-Level Segmentation and Detection:} Segmentation and detection of glomeruli across entire kidney slide images.
\end{enumerate}
Model performance is assessed using the Dice Similarity Coefficient (DSC) and F1-score on the testing dataset.

The KPIs Challenge encompasses a broad spectrum of kidney disease models, including normal and multiple CKD-specific conditions, derived from preclinical rodent models. As a pioneering effort within the MICCAI community, the challenge features an extensive collection of 10,000 normal and diseased glomeruli from over 60 Periodic Acid Schiff (PAS)-stained whole slide images. Each image includes nephrons, with each nephron containing a glomerulus and a small cluster of blood vessels. Participants are tasked with developing algorithms that can accurately segment glomeruli at a pixel level. To the best of our knowledge, this is the first MICCAI challenge exclusively dedicated to segmenting functional units in kidney pathology across diverse CKD models.

\section{Related Works}
    \subsection{Patch-Level Glomerular Segmentation}
    
Patch-level glomerular segmentation has been a focal point in renal pathology, aiming to delineate glomeruli within small, manageable regions of kidney histology images \citep{ginley2019computational}. Early methodologies predominantly utilized classical image processing techniques, including thresholding, edge detection, and morphological operations \citep{electronics9030503, ginley2017automatic}, to identify glomerular structures. While these approaches established foundational workflows, they often encountered challenges due to the intricate morphology of glomeruli and variability in staining methods, leading to inconsistent segmentation outcomes \citep{electronics9030503}.

The advent of deep learning, particularly convolutional neural networks (CNNs), has significantly advanced the field of medical image segmentation \citep{GALLEGO2021101865}. Architectures such as U-Net and its derivatives have been extensively applied to patch-level glomerular segmentation, demonstrating enhanced accuracy and efficiency by learning complex patterns inherent in kidney tissues \citep{Samant2023Glomerulus, GALLEGO2021101865, deng2023omni}.
Despite these advancements, several challenges persist in patch-level segmentation. A primary concern is the lack of global contextual information, as models trained on isolated patches may not effectively capture the spatial relationships and broader tissue architecture present in whole slide images (WSIs) \citep{Wu2023DigitalPathology}. This limitation can lead to inaccuracies, particularly when glomeruli are located near the periphery of patches or exhibit atypical presentations. Additionally, variability in image quality, staining techniques, and the inherent complexity of glomerular structures can adversely affect segmentation performance \citep{Wu2023DigitalPathology}. 
To address these issues, recent research has explored hybrid models that integrate CNNs with transformer-based architectures to enhance both local feature extraction and global context understanding \citep{Liu2024, Yin2024, Wu2023DigitalPathology, deng2024prpseg, deng2024hats, deng2023segment, cuienhancing}. For example, a hybrid CNN-TransXNet \citep{Liu2024} approach has been proposed to improve segmentation accuracy by combining the strengths of CNNs in capturing fine-grained local features with the global contextual understanding provided by transformers.

    \subsection{Whole Slides Image-Level Glomerular Detection and Segmentation}

WSI glomerular detection and segmentation have become pivotal in computational kidnet pathology, enabling comprehensive analysis of kidney tissues at high resolution. Unlike patch-based methods, WSI approaches consider the entire tissue context, facilitating more accurate localization and characterization of glomeruli \citep{BUENO2020105273, tang2024holohisto}. 

Recent advancements have introduced holistic frameworks that integrate detection, segmentation, and lesion characterization within unified pipelines. For instance, the Glo-In-One \citep{GloInOne, GloInOne_v2} toolkit performs comprehensive glomerular quantification from WSIs, streamlining the analysis process for non-technical users. Extension plugin GloFinder \citep{GloFinder}, designed for QuPath, enables single-click automated glomeruli detection across entire WSIs, streamlining the analysis process for non-technical users.

Additionally, deep learning models have been employed to enhance glomerular detection in WSIs, such as modified U-Net architecture \citep{jha2021instance}. These models leverage convolutional neural networks to accurately identify glomeruli, demonstrating significant improvements over traditional methods \citep{jha2021instance, JIANG20211431}. 

Despite these advancements, several challenges persist in WSI glomerular analysis. Processing gigapixel WSIs demands substantial computational resources, necessitating efficient algorithms and hardware \citep{BECKER202065}. Differences in staining protocols, imaging modalities, and tissue preparation can introduce variability, affecting model performance \citep{7243333}. Obtaining pixel-level annotations for WSIs is labor-intensive, often resulting in limited training data for supervised learning models \citep{9973240}. 

To address challenges in WSI glomerular detection and segmentation, researchers have developed innovative strategies. For example, Weighted Circle Fusion (WCF) \citep{WCF} has been proposed to enhance detection precision by fusing outputs from multiple models using confidence-weighted circle representations. This method not only improves accuracy but also reduces false positives by effectively merging overlapping detections. Furthermore, the human-in-the-loop (HITL) approach has significantly improved annotation efficiency, combining machine learning detection with human verification. However, WCF is not without limitations. It demonstrates slightly lower recall compared to methods like Non-Maximum Suppression (NMS), and its reliance on careful parameter tuning may restrict generalizability across different datasets \citep{WCF}.

\section{Challege description}
\subsection{Overview}

The Kidney Pathology Image Segmentation challenge seeks to advance glomeruli segmentation in kidney pathology using deep learning techniques. Participants are tasked with developing precise, pixel-level segmentation algorithms for glomeruli across diverse mouse models, including normal and diseased conditions. The dataset includes whole slide images and patches from four mouse groups: normal, 5/6 nephrectomy (5/6Nx), diabetic nephropathy (DN), and NEP25 models, stained with Periodic acid Schiff (PAS). The challenge emphasizes distinguishing glomeruli from surrounding tissue components under varied preparation scenarios, highlighting the algorithms' versatility and precision.

The competition is divided into two distinct tasks, each addressing specific challenges and emphasizing different aspects of segmentation:

\begin{itemize}
    \item \textbf{Patch-level Segmentation:} This task focuses on the segmentation of glomeruli within specific image patches extracted from the whole slide images. Participants must handle high-resolution data and ensure precise delineation of glomeruli under constrained spatial contexts. This task is ideal for evaluating algorithms on local-level segmentation challenges.
    
    \item \textbf{Whole Slide Image-Level Segmentation:} This task involves segmenting glomeruli across entire kidney slide images, which presents additional challenges such as handling large-scale data, maintaining computational efficiency, and addressing variability across the entire tissue section. It requires algorithms capable of performing robust segmentation on a global scale, adapting to diverse morphological and pathological variations.
\end{itemize}

By addressing these tasks, participants contribute to advancing automated kidney pathology image analysis, demonstrating the adaptability and effectiveness of their algorithms in both localized and comprehensive segmentation scenarios. Participants can choose to tackle either or both tasks, contributing innovative solutions to advance kidney pathology image analysis.

\subsection{Data  description}
\subsubsection{Data overview}

The dataset for the KPIS Challenge includes WSIs derived from four groups of mouse models: 

\begin{itemize}

\item \textbf{Normal group}: Normal mice, sacrificed at the age of 8 weeks.

\item \textbf{5/6Nx group}: Mice underwent 5/6 nephrectomy, sacrificed at 12 weeks after nephrectomy (age of 20 weeks)~\citep{lim2014animal}.

\item \textbf{DN group}: eNOS\textsuperscript{-/-} db/db mice (male or female) sacrificed at the age of 18 weeks~\citep{al2024effects}.

\item \textbf{NEP25 group}: Transgenic mice expressing human CD25 selectively in podocytes, sacrificed at 3 weeks after immunotoxin-induced glomerular injury (age of 11 weeks)~\citep{lim2017tubulointerstitial}.

\end{itemize}

The tissue sections were stained with PAS to visualize morphology, and the stained slides were scanned using the KF-pro-040-Hi scanner (Konfoong Bioinformation Tech Co., Ltd.) at Vanderbilt University Medical Center. Each image encompasses glomeruli, tubules, interstitium, and vasculature. All glomeruli in the digital images were annotated by three experienced pathologists using Qupath software~\citep{bankhead2017qupath}. The WSIs are formatted in pyramid resolution levels of $1, 2, 4, 8, 16, 32, 64$, formatted as TIFF files (.tiff). The image patches are provided at a fixed resolution of $2048 \times 2048$ pixels, formatted as PNG files (.png). 

The dataset is divided into training, validation, and testing sets, with each subset containing annotations for glomeruli segmentation. Table~\ref{tab:dataset_overview} provides an overview of the dataset, detailing the number of WSIs, image resolution, and other characteristics across the subsets.

\begin{table*}[h!]
\centering
\small 
\caption{Summary of the KPIs Challenge Dataset}
\label{tab:dataset_overview}
\begin{tabular}{lcccc|cccc|cccc}
\toprule
\multirow{2}{*}{\textbf{Property}} & \multicolumn{4}{c}{\textbf{Training}} & \multicolumn{4}{c}{\textbf{Validation}} & \multicolumn{4}{c}{\textbf{Testing}} \\ \cline{2-13}
                                   & \textbf{Norm.} & \textbf{5/6Nx} & \textbf{DN} & \textbf{NEP25} & \textbf{Norm.} & \textbf{5/6Nx} & \textbf{DN} & \textbf{NEP25} & \textbf{Norm.} & \textbf{5/6Nx} & \textbf{DN} & \textbf{NEP25} \\ \hline
\textbf{\# WSIs}                  & 5              & 5              & 5           & 10             & 2              & 2              & 2           & 2              & 3              & 3              & 3           & 3              \\ \hline
\textbf{\# Patches}               & 1786           & 648            & 724         & 1760           & 861            & 274            & 299         & 209            & 1166           & 463            & 391         & 285            \\ \hline
\textbf{Resolution (µm/px)}       & 0.11           & 0.11           & 0.11        & 0.24           & 0.11           & 0.11           & 0.11        & 0.25           & 0.11           & 0.11           & 0.11        & 0.25           \\ \hline
\textbf{Optical Mag.}             & 40x            & 40x            & 40x         & 20x            & 40x            & 40x            & 40x         & 40x            & 40x            & 40x            & 40x         & 40x            \\ \hline
\textbf{Digital Mag.}             & 80x            & 80x            & 80x         & 40x            & 80x            & 80x            & 80x         & 40x            & 80x            & 80x            & 80x         & 40x            \\
\bottomrule
\end{tabular}
\end{table*}

\subsection{Challenge Setup}

The KPIs Challenge began with the release of the training dataset on March 30, 2024, hosted on the Synapse platform, which facilitates data sharing. The validation dataset was released on May 1, 2024, allowing participants to evaluate their algorithms locally.

The test dataset remained private, and participants were required to submit Docker containers with their algorithms. These containers were evaluated after the testing phase ended. The evaluation was performed on an Ubuntu 20.04 desktop with the following specifications:

\begin{itemize}
    \item \textbf{CPU}: Intel(R) Xeon(R) Gold 6230R CPU @ 2.10GHz (52 threads)
    \item \textbf{GPU}: NVIDIA RTX A6000 with 64 GB memory
    \item \textbf{RAM}: 64 GB
    \item \textbf{Driver Version}: 535.183.01
    \item \textbf{CUDA Version}: 12.2
    \item \textbf{Docker Version}: 27.0.1
\end{itemize}

The challenge concluded with the announcement of winners on September 15, 2024. Participants were invited to present their methods at the MICCAI 2024 MOVI Workshop on October 10, 2024, and encouraged to contribute to a joint manuscript for dissemination of the challenge outcomes.

\subsection{Metrics and evaluation}
\subsubsection{Choice of metrics}

The KPIs Challenge employs tailored metrics to address the specific objectives of each task:

\begin{itemize}
    \item \textbf{Task 1: Patch-level Glomeruli Segmentation}  
    The Dice Similarity Coefficient (DSC) is used to evaluate the precision of segmentation at the patch level. This metric ensures accurate pixel-level overlap between predicted and ground truth masks, emphasizing the importance of precise segmentation in smaller, localized regions.

    \item \textbf{Task 2a: WSI-level Glomeruli Segmentation}  
   WSI segmentation is significantly more challenging than patch-level segmentation due to the large image size and complexity of WSIs compared to traditional segmentation tasks. These large-scale images often require models to maintain computational efficiency while addressing the diverse spatial and morphological variations of glomeruli. To balance these challenges, the evaluation includes:  
    - \textit{DSC:} Measures the pixel-level accuracy of glomeruli segmentation, ensuring precise delineation.  
    - \textit{F1 Score:} Evaluates the balance between precision and recall, promoting comprehensive glomeruli detection.  
    The final ranking combines these metrics equally, referred to as ``Glo Detection and Segmentation," to reflect both detection comprehensiveness and segmentation precision.

    \item \textbf{Task 2b: WSI-level Glomeruli Detection}  
    For the detection task, the F1 score is calculated independently. This metric prioritizes models that achieve a balance between precision and recall, ensuring accurate detection of as many glomeruli as possible across the entirety of the WSI.
\end{itemize}

These metrics ensure a fair and comprehensive evaluation, addressing the unique challenges posed by patch-level and WSI-level segmentation and detection tasks.
\begin{sidewaystable}
\centering

\caption{This table summarizes and compares the methods proposed by the top 3 performing teams for each challenge task on the leaderboard: Task 1 (Patch-level glomeruli segmentation), Task 2a (WSI-level glomeruli instance segmentation), and Task 2b (WSI-level glomeruli detection). For each method, we highlight methodology and implementation details, including the training and inference strategies.}
\begin{adjustbox}{width=1\textwidth}

\begin{tabular}{|cccc|cccc|ccc|}
\hline
\rowcolor[HTML]{FFFFFF} 
\multicolumn{4}{|c|}{\cellcolor[HTML]{FFFFFF}Methodology}                                                                                                                                                                                                     & \multicolumn{4}{c|}{\cellcolor[HTML]{FFFFFF}Training Strategy}                                                                                                                                                                                                           & \multicolumn{3}{c|}{\cellcolor[HTML]{FFFFFF}Inference Strategy}                                                                                                                                                                                    \\
\rowcolor[HTML]{FFFFFF} 
Rank                                                                                 & Teams         & Methods                                                                       & \begin{tabular}[c]{@{}c@{}}Segmentation \\ networks\end{tabular}       & \begin{tabular}[c]{@{}c@{}}Cropped \\ (Downsample) size\end{tabular}                            & \begin{tabular}[c]{@{}c@{}}Data \\ Augmentations\end{tabular}       & \begin{tabular}[c]{@{}c@{}}Loss \\ function(s)\end{tabular}                       & Optimization & Pre-processing                                                               & Ensembling                                                               & Post-processing                                                                          \\ \hline
\rowcolor[HTML]{FFFFFF} 
\begin{tabular}[c]{@{}c@{}}Task 1, 2a: 1st place\\ (Task 2b: 4th place)\end{tabular} & Capybara      & Ensembling + Stiching                                                         & SegFormer-B5                                                           & 768 $\times$ 768                                                                                & \begin{tabular}[c]{@{}c@{}}Brightness,\\  flips, blur.\end{tabular} & CE + DICE                                                                         & AdamW        & Cropping                                                                     & 3 $\times$ SegFormer-B5                                                  & \begin{tabular}[c]{@{}c@{}}Sum overlapping patches + \\ Softmax + Crop back\end{tabular} \\
\rowcolor[HTML]{EFEFEF} 
Task 1: 2nd place                                                                    & agaldran      & Ensembling                                                                    & FPN/MiT                                                                & \begin{tabular}[c]{@{}c@{}}1024 $\times$ 1024\\ (Downsample)\end{tabular}                       & Custom augm.\footnote{Cropping, intensity scaling, resizing, flipping, rotation, color jittering, sharpness adjustment}                                                        & CE + DICE                                                                         & Nadam        & Downsample + TTA\footnote{Test Time Augmentation: Flipping, Rotation}                                                             & 5 $\times$ FPN/MiT                                                       & Upsample                                                                                 \\
\rowcolor[HTML]{FFFFFF} 
Task 1: 3rd place                                                                    & Aira Matrix & Ensembling                                                                    & SegNeXt                                                                & \begin{tabular}[c]{@{}c@{}}1024 $\times$1024, \\ 512 $\times$512\end{tabular}                   & Custom augm.\footnote{Flipping, sptail scaling, shifting, rotations, color jittering, gamma correction}                                                        & \begin{tabular}[c]{@{}c@{}}CE + IoU loss + \\ Entropy regularization\end{tabular} & AdamW        & Cropping                                                                     & 2 $\times$ SegNeXt                                                       & \begin{tabular}[c]{@{}c@{}}Weighted average + \\ Argmax\end{tabular}                     \\
\rowcolor[HTML]{EFEFEF} 
\begin{tabular}[c]{@{}c@{}}Task 2a: 2nd place\\ (Task 2b: 3rd place)\end{tabular}    & Zhijian Life  & Ensembling  + Stiching                                                        & \begin{tabular}[c]{@{}c@{}}ResNet101 +\\ Swin Transformer\end{tabular} & \begin{tabular}[c]{@{}c@{}}4,096 $\times$ 4,096 (80x)\\ 2,048 $\times$ 2,048 (40x)\end{tabular} &                                                                     &                                                                                   &              & \begin{tabular}[c]{@{}c@{}}OTSU + Cropping + \\ Filtering noise\end{tabular} & \begin{tabular}[c]{@{}c@{}}ResNet101 +\\ Swin Transformer\end{tabular}   & \begin{tabular}[c]{@{}c@{}}Edge patch: Paste\\ Center patch: Crop, Paste\end{tabular}    \\
\rowcolor[HTML]{FFFFFF} 
\begin{tabular}[c]{@{}c@{}}Task 2a: 2nd place\\ (Task 2b: 5th place)\end{tabular}    & Deep Bio       & Ensembling + Stiching                                                         & Hybrid.\footnote{Encoder: ResNet50 + ViT-B/16; Decoder: SETR}                                                                & 1024 $\times$ 1024                                                                              & Custom augm.\footnote{Flipping, rotations, resizing, color jittering, gaussian filtering, replacing 10\% image in mini batch with ImageNet21k, Mixup, CutMix}                                                        & DICE                                                                              & AdamW        & Cropping                                                                     & Hybrid. (4 seeds)                                                        & Average + Thresholding                                                                   \\
\rowcolor[HTML]{EFEFEF} 
\begin{tabular}[c]{@{}c@{}}Task 2b: 1st place\\ (Task 2a: 4th place)\end{tabular}    & salt\_fish    & \begin{tabular}[c]{@{}c@{}}SAM (LoRA) + \\ Ensembling + Stiching\end{tabular} & ViT-H (SAM)                                                            & 512 $\times$ 512                                                                                & Custom augm.\footnote{Flipping, spatial scaling, rotation, gaussian noise and filtering, contrast adjustment}                                                                    & CE + DICE                                                                         & AdamW        & \begin{tabular}[c]{@{}c@{}}Cropping + \\ Add extra dim.\end{tabular}         & ViT-H (4 seeds)                                                          & \begin{tabular}[c]{@{}c@{}}Sigmoid + Thresholding +\\ Remove extra dim.\end{tabular}     \\
\rowcolor[HTML]{FFFFFF} 
\begin{tabular}[c]{@{}c@{}}Task 2b: 2nd place\\ (Task 2a: 5th place)\end{tabular}    & CVAILAB       & Ensembling + Stiching                                                         & nnU-Net                                                                &                                                                                                 & nnU-Net augm.                                                       &                                                                                   &              & \begin{tabular}[c]{@{}c@{}}Cropping + \\ Edge removal\end{tabular}           & \begin{tabular}[c]{@{}c@{}}nnU-Net + \\ DeepLabV3 + U-Net++\end{tabular} & Center patch: stich + resize                                                             \\ \hline

\end{tabular}

\label{tab:methods_summary}

\end{adjustbox}

\end{sidewaystable}

\section{Participating methods}
In this section, we provide a summary of the participating methods for each challenge task by selecting the top 3 performing teams as representatives and succinctly describing their approaches in the following section. Table \ref{tab:methods_summary} provides a brief comparison of the proposed approaches, highlighting their methodology and implementation details, including the training and inference strategies.

\subsection{Task 1: Patch-level Glomeruli Segmentation}

\noindent\textit{\textbf{Capybara (1st place, Cap)}}. The proposed glomerulus segmentation pipeline is based on patch-level model ensembling and stitching. A crop size of 768 $\times$ 768 was used during both model training and inference. To account for the various CKD in the challenge data, the training data was stratified into three folds based on disease model and WSI name. Three SegFormer-B5 \citep{xie2021segformer} models were trained on the three folds and combined into one unified model. During inference, for each image patch of size 2,048 $\times$ 2,048, a window size of 768 × 768 with a stride size of 576 (75\% of the window size) were used. Predictions from the three models are averaged. To enhance detection coverage, the pipeline applies stitching to overlapping image patches based on information such as their coordinates and WSI names. Specifically, for overlapping patches, the raw prediction values in the intersecting areas are summed. The stitched prediction map from all image patches of the same WSI is normalized using the softmax function. Finally, the results are cropped back to the original test size (2,048 $\times$ 2,048) for submission. Code is available at \url{https://github.com/huuquan1994/wsi_glomerulus_seg}.

\noindent\textit{\textbf{agaldran (2nd place, Galdran et al.)}}. The proposed method focuses on segmenting glomeruli from smaller, predefined patches of the WSI. The input image patches were resized to 1,024 $\times$ 1,024 for both model training and inference. The segmentation model is based on an encoder-decoder architecture, with the ImageNet-pretrained Feature Pyramid Network (FPN) \citep{lin2017feature}  serving as the encoder and the Mix Vision Transformer (MiT) \citep{xie2021segformer} as the decoder. FPN employs a top-down architecture with lateral connections to extract multi-scale features. The MiT decoder, based on a transformer architecture, processes and integrates the features extracted by the encoder into a segmentation map, effectively preserving both global and local context. The model was trained using a five-fold cross-validation scheme. During inference, Test-Time Augmentation (TTA) was applied, which involved flipping each patch horizontally and/or vertically. The final segmentation result was obtained by averaging the predictions from the five cross-validation models. Code is available at \url{https://github.com/agaldran/kpis/tree/main}.

\noindent\textit{\textbf{Aira Matrix (3rd place, A. Kumar et al.)}}. This team introduced an ensemble SegNeXt-based method \citep{guo2022segnext} for patch-level glomeruli segmentation, where two models were trained using image patches of different cropped sizes. First, input image patches are randomly cropped to sizes of 512 $\times$ 512 and 1,024 $\times$ 1,024 to train two separate SegNeXt \citep{guo2022segnext} models, both using ConvNeXt Tiny \citep{liu2022convnet} as the backbone. During training, the ImageNet-pretrained weights are used to initialize the encoder network. An Entropy Regularized Segmentation (ERS) loss is introduced, which combines cross-entropy loss and IoU loss with an additional entropy regularization term. During inference, the two SegNeXt models are applied to preprocessed image patches with different sizes (512 $\times$ 512 and 1,024 $\times$ 1,024). The outputs from both models are then merged using a weighted average, and the combined model prediction is passed through the Argmax function.

\subsection{Task 2a: WSI-level Glomeruli Segmentation}

\noindent\textit{\textbf{Capybara (1st place, Cap)}}. The same pipeline in Task 1 (patch-level segmentation) was also used in this WSI-level glomerulus segmentation task. For WSI-level data, a sliding window with a crop size of 2,048 $\times$ 2,048 and a stride size of 1,024 $\times$ 1,024 was used to extract image patches for inference. During the inference stage, the same ensemble model used in Task 1 was applied to generate patch-level predictions. Subsequently, the proposed stitching strategy was employed to merge the predictions from all image patches into a unified prediction map, maintaining the original size of the corresponding WSI. Specifically, for overlapping image patches, the raw prediction values in the intersecting areas are summed. Finally, the stitched prediction map is normalized using the softmax function. This approach improves detection coverage, particularly when glomeruli are located near the borders of the patch images. Code is available at \url{https://github.com/huuquan1994/wsi_glomerulus_seg}.

\noindent\textit{\textbf{Zhijian Life (2nd place, Cai et al.)}}. This team implemented a robust deep learning method for WSI-level diseased glomeruli segmentation, built on top of a conventional pipeline. First, conventional OTSU thresholding is applied to obtain tissue regions. Small areas (less than 2,000 pixels) are filtered out to reduce noise for subsequent analysis. During segmentation model training, both the WSI-level image and mask go through the Overlapping Patch extraction block to prepare the data for training, with 4,096 $\times$ 4,096 (80x.magnification) and  2,048 $\times$ 2,048 (40x.magnification). ResNet101 \citep{he2016deep} and Swin Transformer \citep{liu2021swin} are combined with the UPerNet \citep{xiao2018unified} framework to train on the extracted image patches. At the inference stage, the patch-level segmentation model (e.g., ResNet101 and Swin Transformer) predictions are aggregated to generate the WSI-level prediction masks. First, an all-black image ($\vartheta$) with the same resolution as the WSI is generated. Then, two patch aggregation strategies are implemented. if a patch is located at the edge (top, right, left, or bottom) of the tissue, it is pasted directly into $\vartheta$ (using Swin Transformer prediction); otherwise, only the central area of the patch is cropped and pasted into $\vartheta$ (using ResNet101 prediction).

\noindent\textit{\textbf{Deep Bio (2nd place, Lee et al.)}}. The proposed method employs a hybrid model architecture to achieve higher performance while handling insufficient data. To train and infer with the segmentation model, the WSIs are tiled into 1,024 $\times$ 1,024 image patches with 50\% overlap both horizontally and vertically. The hybrid segmentation encoder combines low-level image features extracted by ResNet50 \citep{he2016deep} with the global attention mechanism of ViT-B \citep{dosovitskiy2020image}. The segmentation decider is SETR \citep{zheng2021rethinking} segmentation head. During the training stage, both the CNN and transformer models employ pretrained weights from ImageNet. Additionally, to address the challenge of limited training data, strong data augmentation is applied. In addition to normal data augmentation, 10\% of images in the mini-batch are replaced with images from ImageNet 21K. Advanced data augmentations, such as MixUp \citep{zhang2017mixup} and CutMix \citep{yun2019cutmix}, are also utilized. To enhance robustness, four ensemble models with different random seeds are trained. During the inference stage, patch-level segmentation results are merged, with the values in the overlapping areas averaged. The heatmaps from the four ensemble models are averaged and a threshold of 0.5 is applied.

\subsection{Task 2b: WSI-level Glomeruli Detection}
\noindent\textit{\textbf{salt\_fish (1st place, Chen et al.)}}. The proposed method utilizes the Segment Anything Model \citep{kirillov2023segment} as the image encoder with Low-Rank Adaptation (LoRA) \citep{hu2021lora} fine-tuning. First, at the preprocessing stage, the input data are normalized, and an additional dimension is added to the 2D labels. During training, the input data are randomly cropped into 512 $\times$ 512 patches. SAM (ViT-H) is used as the encoder and fine-tuned with LoRA using four different initializations. During inference, a sliding window with a patch size of 512 $\times$ 512 is utilized. The network outputs undergo a sigmoid activation function and are thresholded at 0.5. Then, the added dimension is removed, and objects smaller than 10,000 pixels are deleted. Finally, the segmentation outputs from the four ensemble SAM models are merged.

\noindent\textit{\textbf{CVAILAB (2nd place, Hou et al.)}}.
This team implements a WSI-level glomeruli segmentation pipeline with adaptive cropping and an nnU-Net \citep{isensee2021nnu}-based architecture. To handle edge effects in WSI glomeruli segmentation, image patches are cropped with overlaps, and edge removal is performed to prevent glomeruli fragmentation at patch edges. Two modified U-Net \citep{ronneberger2015u} architectures, U-Net++ \citep{zhou2018unet++} and U-Net3+ \citep{huang2020unet}, are implemented based on nnU-Net to allow denser network connections and improved model generalizability. During inference, ensemble learning is applied to adaptively cropped image patches for robust and generalized mask results. Central patch masks are stitched back into the WSI-level mask using recorded coordinates.

\noindent\textit{\textbf{Zhijian Life (3rd place, Cai et al.)}}. This team implemented a robust deep learning method for WSI-level diseased glomeruli segmentation, built on top of a conventional pipeline. First, conventional OTSU thresholding is applied to obtain tissue regions. Small areas (less than 2,000 pixels) are filtered out to reduce noise for subsequent analysis. During segmentation model training, both the WSI-level image and mask go through the Overlapping Patch extraction block to prepare the data for training, with 4,096 $\times$ 4,096 (80x.magnification) and  2,048 $\times$ 2,048 (40x.magnification). ResNet101 \citep{he2016deep} and Swin Transformer \citep{liu2021swin} are combined with the UPerNet \citep{xiao2018unified} framework to train on the extracted image patches. At the inference stage, the patch-level segmentation model (e.g., ResNet101 and Swin Transformer) predictions are aggregated to generate the WSI-level prediction masks. First, an all-black image ($\vartheta$) with the same resolution as the WSI is generated. Then, two patch aggregation strategies are implemented. if a patch is located at the edge (top, right, left, or bottom) of the tissue, it is pasted directly into $\vartheta$ (using Swin Transformer prediction); otherwise, only the central area of the patch is cropped and pasted into $\vartheta$ (using ResNet101 prediction).

\begin{table}[t]
\caption{Task 1: Patch-level Diseased Glomeruli Segmentation Leaderboard. The submissions of the top 10 teams are presented, along with their Dice scores, shown as Mean$\pm$SD.}
\centering
\scriptsize
\setlength{\tabcolsep}{4mm}
\renewcommand\arraystretch{1}
\begin{tabular}{c|c|c}
\toprule
Rank & Team & Dice Score \\
\midrule
1 & Capybara & 94.51$\pm$6.71 \\
2 & agaldran & 94.28$\pm$5.81 \\
3 & Aira Matrix & 94.28$\pm$5.89 \\
4 & mafi95 & 94.16$\pm$6.86\\
5 & Deep Bio & 93.64$\pm$6.00 \\
6 & Birmingham Vanguard AI Lab & 93.62$\pm$4.52 \\
7 & p53 & 93.23$\pm$7.65 \\
8 & Bio-Totem & 93.08$\pm$7.24 \\
9 & CVAILAB & 92.79$\pm$8.28 \\
10 & NJUST Vision Group & 92.70$\pm$7.73 \\
\bottomrule
\end{tabular}
\label{tab:Task1Dice}
\end{table}

\begin{table}[t]
\caption{Task 2a: WSI-level Diseased Glomeruli Instance Segmentation Leaderboard. The top 10 teams are provided, with the overall rank based on the F1 score (detection) and Dice score (segmentation) performance. The values for both metrics are presented as Mean$\pm$SD.}
\centering
\scriptsize
\setlength{\tabcolsep}{1.5mm}
\renewcommand\arraystretch{1}
\begin{tabular}{cccccc}
\toprule
Overall Rank & Team & Dice Score & F1 Score & Dice Rank & F1 Rank \\
\midrule
1 & Capybara & 94.64$\pm$0.89 & 88.63$\pm$5.80 & 1 & 4 \\
2 & Zhijian Life & 93.36$\pm$1.99  & 89.99$\pm$4.31 & 4 & 3 \\
2 & Deep Bio & 94.48$\pm$1.28 & 87.96$\pm$6.08 & 2 & 5 \\
4 & salt\_fish & 89.81$\pm$2.18 & 91.33$\pm$3.84 & 8 & 1 \\
5 & CVAILAB & 89.37$\pm$3.45 & 90.88$\pm$4.35 & 9 & 2 \\
5 & p53 & 92.74$\pm$1.62 & 86.81$\pm$5.42 & 5 & 6 \\
7 & BozekLab & 91.35$\pm$3.05 & 86.48$\pm$7.96 & 6 & 7 \\
7 & mafi95 & 93.68$\pm$1.14 & 74.30$\pm$17.18 & 3 & 10 \\
9 & TeamTiger & 89.83$\pm$5.26 & 84.96$\pm$7.81 & 7 & 9 \\
10 & XJTU & 84.47$\pm$5.90 & 85.84$\pm$7.14 & 10 & 8 \\
\bottomrule
\end{tabular}
\label{tab:Task2Leaderboard}
\end{table}

\begin{table}[t]
\caption{Task 2b: WSI-level Diseased Glomeruli Detection Leaderboard. The submissions of the top 10 teams are presented, along with their F1 scores, shown as Mean$\pm$SD.}
\centering
\scriptsize
\setlength{\tabcolsep}{4mm}
\renewcommand\arraystretch{1}
\begin{tabular}{c|c|c}
\toprule
Rank & Team & F1 Score \\
\midrule
1 & salt\_fish & 91.33$\pm$3.84 \\
2 & CVAILAB & 90.88$\pm$4.35\\
3 & Zhijian Life & 89.99$\pm$4.31\\
4 & Capybara & 88.63$\pm$5.80\\
5 & Deep Bio & 87.96$\pm$6.08\\
6 & p53 & 86.81$\pm$5.42\\
7 & BozekLab & 86.48$\pm$7.96\\
8 & XJTU & 85.84$\pm$7.14\\
9 & TeamTiger & 84.96$\pm$7.81\\
10 & mafi95 & 74.30$\pm$17.18\\
\bottomrule
\end{tabular}
\label{tab:Task2Detection}
\end{table}

\begin{figure*}[t]
\begin{center}
\includegraphics[width=0.9\linewidth]{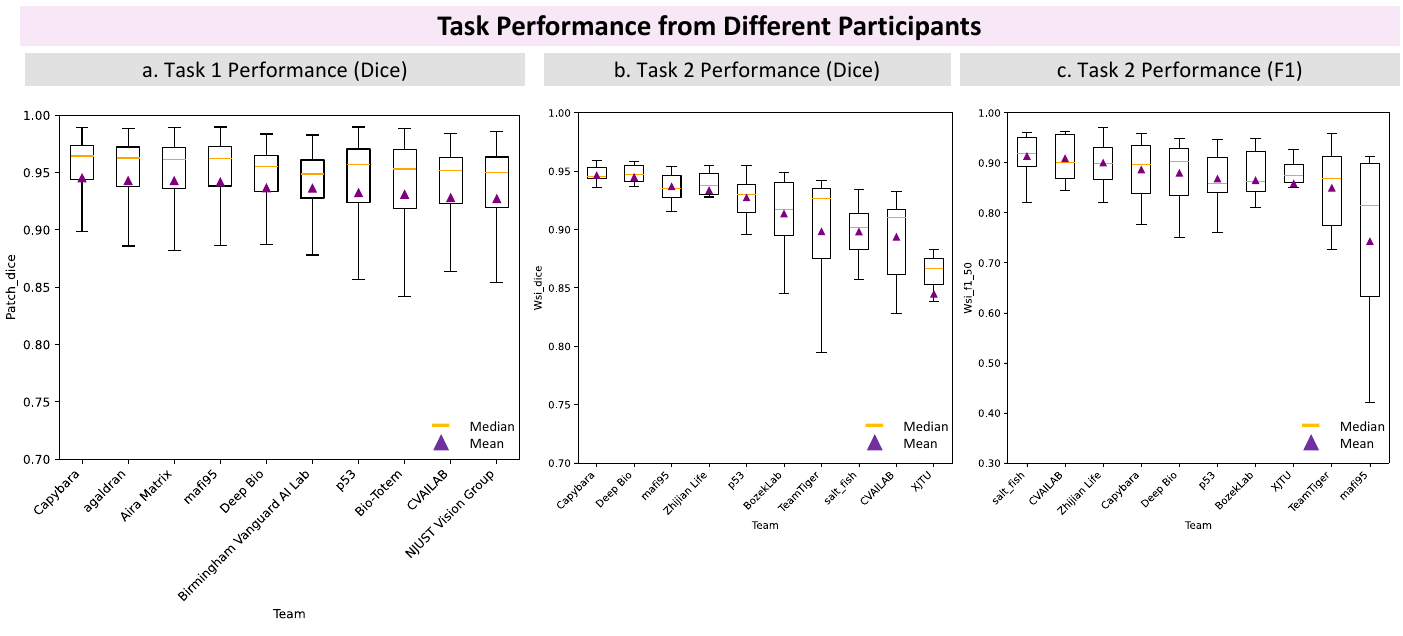}
\end{center}
\caption{\textbf{Box plots of quantitative results on different tasks from participants.} This figure shows the box plots of the quantitative performance of participants on two tasks. Results from the top 10 teams are reported. Dice similarity coefficient scores are provided for Task 1, while both Dice similarity coefficient scores and F1 scores are presented for Task 2.}

\label{fig:boxplots}
\end{figure*}

\begin{figure*}[t]
\begin{center}
\includegraphics[width=1\linewidth]{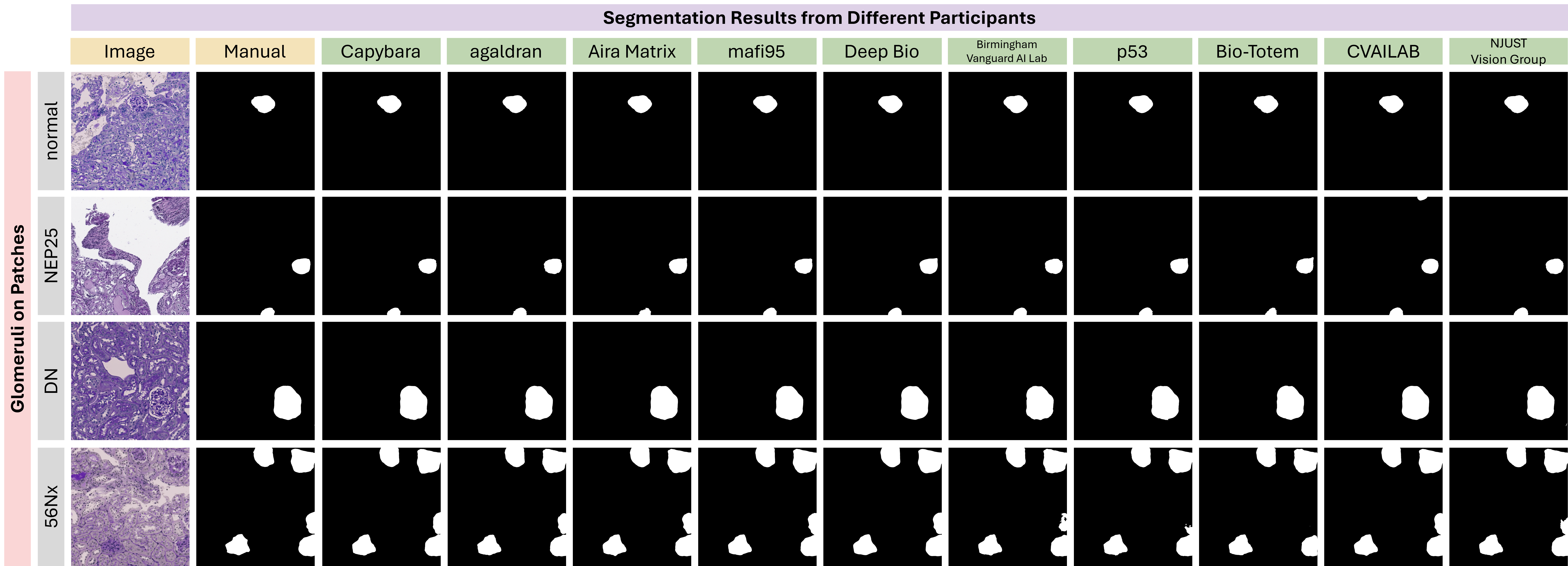}
\end{center}
\caption{\textbf{Patch-level quantitative results} - The segmentation results from different participants' methods on glomeruli under various experimental conditions (normal, NEP25, DN, and 5/6Nx).}
\label{fig:task1_quantitative}
\end{figure*}

\begin{figure*}[t]
\begin{center}
\includegraphics[width=1\linewidth]{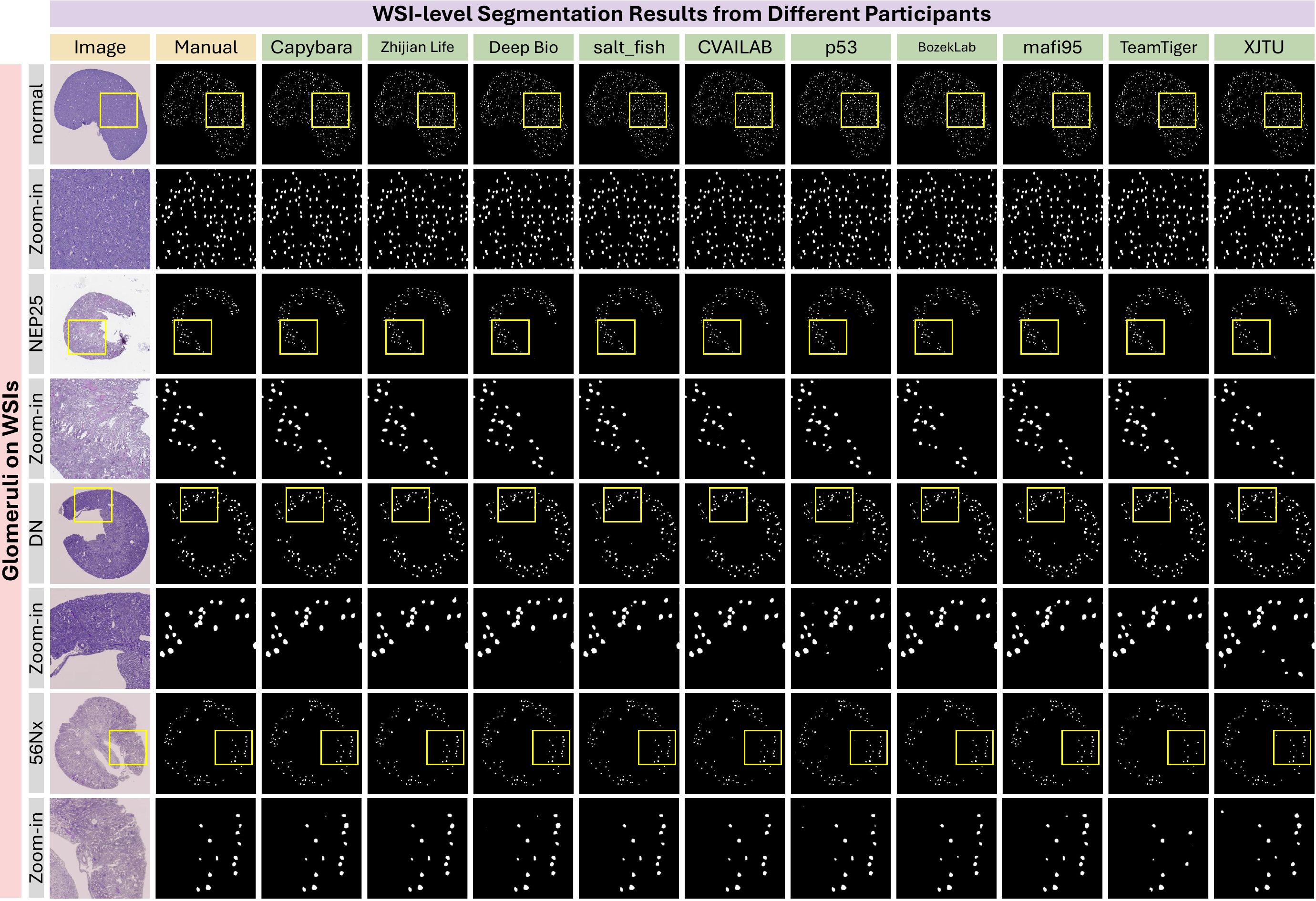}
\end{center}
\caption{\textbf{WSI-level quantitative results} - This figure shows the segmentation results from different participants' methods on glomeruli under various experimental conditions (normal, NEP25, DN, and 5/6Nx). The yellow bounding box highlights the zoomed-in region on the WSI, with the zoomed-in images below showing the corresponding area.}
\label{fig:task2_quantitative}
\end{figure*}

\section{Results}

Participants were required to submit their inference models in the form of a Docker container by August 1, 2024. These submissions were evaluated using our withheld test set, and the final results and leaderboard for the challenge were announced on September 15, 2024. A total of 24 valid submissions were received for Task 1 (Patch-level segmentation) and 15 valid submissions for Task 2 (WSI-level segmentation). The complete leaderboards for all participants can be found at~\footnote{https://sites.google.com/view/kpis2024/leaderboard}.

This section presents the overall segmentation performance for Task 1 and Task 2. Specifically, it includes both qualitative and quantitative results for the top 10 teams on the leaderboard for each task. The complete benchmarking details, derived from an open-source toolkit for analyzing and visualizing challenges~\citep{wiesenfarth2021methods}, are provided in \S\textbf{Supplementary Material}~\ref{sec:supplementary}.

\subsection{Patch-level Segmentation Performance}
This section reports the top 10 teams on the patch-level glomeruli segmentation leaderboard. Table\ref{tab:Task1Dice} presents their ranked Dice scores with mean $\pm$ standard deviation (SD), while Fig.\ref{tab:Task1Dice}a visually illustrates the Dice score distribution as a boxplot. Both the table and figure show that the top 10 teams performed well, with closely aligned mean and median Dice scores, indicating high consistency across their models. \textbf{Capybara} leads with a Dice score of 94.51 $\pm$ 6.71, followed by \textbf{agaldran} and \textbf{Aira Matrix} at 94.28 $\pm$ 5.81 and 94.28 $\pm$ 5.89, respectively. These top 3 teams, summarized as representatives in the previous section, exhibit relatively high mean Dice scores and lower standard deviations compared to lower-ranked teams, which show greater variability, as indicated by SD values exceeding 7.24. Fig.\ref{fig:boxplots}a confirms this trend, with most teams demonstrating small variability, while others show more inconsistency. Fig.\ref{fig:task1_quantitative} provides qualitative patch-level segmentation predictions alongside ground truth, demonstrating stable model performance across all listed teams for four types of glomeruli samples. Top-performing teams (left columns) handle boundary segmentation more effectively than lower-ranked teams (right columns), as seen in the 5/6Nx and NEP25 rows. A challenge in this task is the large size of the given image patches (2,048 $\times$ 2,048), where patch cropping could compromise accuracy at the boundaries, particularly when merging predictions. The stronger performance of the top teams (e.g., \textbf{Capybara} and \textbf{Aira Matrix}) suggests they have implemented effective post-processing strategies to address this challenge.

\subsection{WSI-level Segmentation Performance}
The transition from patch-level to slide-level analysis introduced new challenges, particularly in handling large-scale gigapixel images, which demand significant computational resources. This necessitates image cropping to manage these vast images, followed by addressing boundary effects when merging predictions from cropped patches back to the WSI level. In this section, we present the WSI-level glomeruli instance segmentation results for Task 2a, evaluated using Dice and F1 scores. Additionally, we provide the WSI-level detection rankings for Task 2b, based solely on the F1 score.

\subsubsection{Instance Segmentation}
We also report the top 10 teams on the WSI-level glomeruli instance segmentation leaderboard (Table \ref{tab:Task2Leaderboard}). Both Dice and F1 scores are used to calculate the overall rank for this task. Fig.\ref{fig:boxplots}b and c visually depict the distributions of Dice and F1 scores as boxplots.  Fig.\ref{fig:task2_quantitative} shows a zoomed-in visualization of WSI-level segmentation masks alongside ground truth, where the top-performing teams are ranked from left to right across the columns.

Overall, compared to the patch-level task results from Table \ref{tab:Task2Leaderboard} and Fig.\ref{fig:boxplots}a, the WSI-level instance segmentation task is more challenging, as indicated by the broader range of both segmentation (Dice) and detection (F1) metrics in the boxplots. \textbf{Capybara} emerges as the top performer, leading in Dice with a score of 94.64 $\pm$ 0.89 and ranking 4th in F1 with a score of 88.63 $\pm$ 5.80. \textbf{Zhijian Life} and \textbf{Deep Bio} follow closely, sharing the second rank, with \textbf{Zhijian Life} excelling in detection (F1) and \textbf{Deep Bio} leading in segmentation (Dice). The top teams consistently demonstrate higher mean Dice and F1 scores, with relatively smaller SD compared to lower-ranked teams. Among all teams, \textbf{Capybara} has the highest mean Dice score and the lowest SD of 0.89, making it the only team with a Dice SD below 1. \textbf{Zhijian Life} achieves a Dice score of 93.36 $\pm$ 1.99 and an F1 score of 89.99 $\pm$ 4.31, while \textbf{Deep Bio} scores 94.48 $\pm$ 1.28 in Dice and 87.96 $\pm$ 6.08 in F1. These smaller SD values indicate that the top teams' models are more stable, with less variation in their performance across different WSI samples. In contrast, lower-ranked teams show larger SD values, with Dice SD exceeding 5 and F1 SD reaching as high as over 17. For both evaluation metrics, the difference between the highest and lowest mean scores can exceed 10, highlighting the inconsistency and variability in WSI-level segmentation and detection tasks. To visually confirm, an example of the zoomed-in visualization of a WSI-level segmentation mask in Fig.\ref{fig:task2_quantitative} shows the segmentation performance among different participants. As shown, for the glomeruli types NEP25, DN, and 5/6Nx, the difference in segmentation accuracy is more pronounced between top-performing (left columns) and lower-ranked teams (right columns).

\subsubsection{Detection}
In this task, the F1 score is the only metric considered for evaluating WSI-level prediction masks among participants. Table \ref{tab:Task2Detection} reports the top 10 teams on the WSI-level glomeruli detection leaderboard, ranked by F1 scores with mean $\pm$ SD. Fig. \ref{fig:boxplots}c shows the F1 score distribution as a boxplot. Similarly, Fig.\ref{fig:task2_quantitative} visually compares the WSI-level glomeruli detection performance with zoomed-in regions, where the top-performing teams are ranked from left to right across the columns. 

Among the participants, \textbf{salt\_fish} emerges as the top performer, achieving an F1 score of 91.33 $\pm$ 3.84. \textbf{CVAILAB} and \textbf{Zhijian Life} follow closely, with F1 scores of 90.88 $\pm$ 4.35 and 89.99 $\pm$ 4.31, respectively. As observed earlier, the WSI-level task proposes new challenges and leads to more varied team performance. However, the top-performing teams in detection tasks generally have better rank in instance segmentation Task 2a, which can be also derived from Table.\ref{tab:methods_summary}. Similarly, these teams exhibit high F1 detection scores with relatively low variability in their results, as indicated by smaller SD. In contrast, teams ranked lower demonstrate more variability, with SD exceeding 7 for most, which indicates the importance of using rigorous pre-processing and post-processing methods (e.g., \textbf{Zhijian Life, CVAILAB}), ensembling methods, or even foundation models (e.g., \textbf{salt\_fish}) in the slide-wise task.

\section{Discussion}
The performance of the leading participants in Task 1 was remarkably similar, with Dice scores exceeding 94\% and standard deviations below 7, indicating that most methods performed at a high level with only subtle differences in segmentation accuracy (Table~\ref{tab:Task1Dice}, Fig.\ref{tab:Task1Dice}). However, in Task 2, greater variation in performance was observed across teams, as reflected in the broader range of Dice and F1 scores (Table\ref{tab:Task2Leaderboard}, Table~\ref{tab:Task2Detection}, Fig.~\ref{fig:boxplots}.b and c). This variability suggests that while patch-based segmentation models generalize well, whole slide image (WSI)-level detection and segmentation are significantly influenced by ensembling strategies and post-processing techniques, underscoring the practical differences in the robustness and adaptability of leading models.

The results of Task 1 demonstrate strong segmentation performance across participants, with subtle variations among the top-performing methods. The consistently high Dice scores suggest that the dataset and task were effective in isolating patch-level features critical to glomerular segmentation. The leading methods, such as \textbf{Capybara} and \textbf{Aira Matrix} leveraged ensemble techniques and multi-scale processing, which underscores the importance of integrating local patch-level detail with robust model ensembling. Future iterations of this challenge might benefit from introducing more variability in patch conditions to further differentiate model capabilities, particularly in handling atypical or edge-case glomeruli.

The transition from patch-level to slide-level analysis presented new challenges, as evidenced by the broader range of performance metrics, particularly in F1 scores. The variability in slide-wise detection highlights the importance of ensembling and post-processing strategies to maintain precision and recall at larger scales. While the \textbf{Capybara} model achieved the highest segmentation scores, other participants demonstrated competitive detection accuracy, reflecting a trade-off between segmentation precision and detection comprehensiveness. The inclusion of large-scale data handling and diverse morphological conditions in Task 2 provided a robust testbed for evaluating scalability and adaptability.

A recurring trend among high-performing methods was the athe use of transformers or hybrid architectures, combining CNNs with transformers, for constructing the image encoder. This approach effectively captured both local and global features, a critical requirement for tasks spanning diverse spatial scales. The application of TTA and advanced data augmentation techniques, such as CutMix and MixUp, further enhanced model robustness, suggesting that these strategies are indispensable for generalizing across heterogeneous and imbalanced datasets. Additionally, to handle boundary effects while merging patch-wise predictions back into the slide-wise mask, model ensembling is used along with stitching. There are two main categories of stitching methods. The first involves merging patches straightforwardly and then applying normalization or thresholding to refine the merged masks (\textbf{Capybara}, \textbf{Aira Matrix}, \textbf{Deep Bio}, and \textbf{salt\_fish}). The second is the crop-and-paste method, which, depending on the data and task, handles image patches at the edges of glomeruli tissue or at the center of the glomeruli tissue differently (e.g., \textbf{Zhijian Life} and \textbf{CAVILAB}).

\noindent\textbf{Limitations and Future Directions:} While the dataset included a variety of CKD models, future challenges could benefit from incorporating more extensive interspecies data and additional staining techniques to simulate broader clinical scenarios. Expanding the dataset to include both normal tissue objects and lesion objects could enhance the clinical value of the models. Currently, slide-wise glomeruli detection and segmentation rely heavily on pretrained weights and ensembling, which overlook global morphology and multi-scale patterns inherent to the unique data type of pathology. Encouraging solutions with fewer preprocessing requirements and incorporating designs that leverage the unique properties of pathological data could help address these limitations.

\section{Conclusion}
The Kidney Pathology Image Segmentation Challenge was organized to address existing limitations in glomerular segmentation research by expanding the dataset to include whole kidney sections from diseased rodent models. This competition aimed to advance state-of-the-art segmentation methods for glomerular identification across diverse CKD models, challenging participants to develop robust algorithms capable of pixel-level segmentation under varying tissue conditions and preparation scenarios. By addressing challenges such as variations in glomerular size, shape, and structural integrity, the KPIs Challenge not only fostered innovation but also demonstrated the adaptability and precision of modern segmentation approaches, paving the way for improved tools in kidney pathology research.

\section{Acknowledgements}
This research was supported by NIH R01DK135597 (Huo), DoD HT9425-23-1-0003 (HCY), and KPMP Glue Grant. This work was also supported by Vanderbilt Seed Success Grant, Vanderbilt Discovery Grant, and VISE Seed Grant. This project was supported by The Leona M. and Harry B. Helmsley Charitable Trust grant G-1903-03793 and G-2103-05128. This research was also supported by NIH grants R01EB033385, R01DK132338, REB017230, R01MH125931, and NSF 2040462. We extend gratitude to NVIDIA for their support by means of the NVIDIA hardware grant. This work was also supported by NSF NAIRR Pilot Award NAIRR240055.

\bibliographystyle{model2-names.bst}\biboptions{authoryear}
\bibliography{refs}

\appendix

\section*{Supplementary Materials}
\label{sec:supplementary}
The supplementary materials provide an in-depth benchmarking analysis of diseased glomeruli segmentation across two tasks. Task 1 focuses on patch-level diseased glomeruli segmentation, assessing ten algorithms over 2,305 cases. The ranking methodology relies on the mean metric values, and multiple visualization techniques, such as boxplots, podium plots, and ranking heatmaps, illustrate performance differences among algorithms. Stability analyses using bootstrap sampling, significance maps, and ranking robustness comparisons further validate the results. Task 2 extends the evaluation to WSI-level diseased glomeruli instance segmentation, comparing algorithm performance across different ranking criteria, including Dice and F1 scores. This study examines 10 algorithms over 12 cases and integrates cross-task insights, exploring algorithm variability through dendrogram clustering and ranking robustness across different evaluation methods. These supplementary results offer valuable insights into algorithmic performance, reliability, and comparative effectiveness in glomeruli segmentation tasks.

\clearpage


\section*{Supplementary material}

# Benchmarking report for Task2 WSI-level Diseased Glomeruli Instance Segmentation

## created by challengeR v1.0.5

## 07 February, 2025

This document presents a systematic report on the benchmark study "Task2 WSI-level Diseased Glomeruli Instance Segmentation". Input data comprises raw metric values for all algorithms and cases. Generated plots are:

- Visualization of assessment data: Dot- and boxplots, podium plots and ranking heatmaps
- Visualization of ranking stability: Blob plots, violin plots and significance maps, line plots
- Visualization of cross-task insights: Blob plots, stacked frequency plots, dendrograms

Details can be found in Wiesenfarth et al. (2021).

# 1 Rankings

Algorithms within a task are ranked according to the following ranking scheme:

*aggregate using function ("mean") then rank*

Ranking for each task:

Task2_dice : The analysis is based on 10 algorithms and 12 cases. 0 missing cases have been found in the data set.

|              | MetricValue__mean | rank |
|--------------|-------------------|------|
| Capybara     | 0.9463778         | 1    |
| Deep Bio     | 0.9448269         | 2    |
| mafi95       | 0.9368086         | 3    |
| Zhijian Life | 0.9336372         | 4    |
| p53          | 0.9273659         | 5    |
| BozekLab     | 0.9135242         | 6    |
| TeamTiger    | 0.8982856         | 7    |
| salt__fish   | 0.8981474         | 8    |
| CVAILAB      | 0.8936657         | 9    |
| XJTU         | 0.8446779         | 10   |

Task2_F1 : The analysis is based on 10 algorithms and 12 cases. 0 missing cases have been found in the data set.

|            | MetricValue_mean | rank |
|------------|------------------|------|
| salt_fish  | 0.9133389        | 1    |
| CVAILAB    | 0.9088149        | 2    |
| Zhijian Life | 0.8998728      | 3    |
| Capybara   | 0.8862561        | 4    |
| Deep Bio   | 0.8795688        | 5    |
| p53        | 0.8680942        | 6    |
| BozekLab   | 0.8648493        | 7    |
| XJTU       | 0.8583894        | 8    |
| TeamTiger  | 0.8496339        | 9    |
| mafi95     | 0.7430213        | 10   |

Consensus ranking across tasks according to chosen method "Borda":

|             | value | rank |
|-------------|-------|------|
| Capybara    | 1.0   | 1    |
| Deep Bio    | 2.5   | 2    |
| Zhijian Life | 2.5  | 2    |
| salt_fish   | 4.0   | 4    |
| CVAILAB     | 5.5   | 5    |
| p53         | 5.5   | 5    |
| BozekLab    | 7.5   | 7    |
| mafi95      | 7.5   | 7    |
| TeamTiger   | 9.0   | 9    |
| XJTU        | 10.0  | 10   |

# 2 Visualization of raw assessment data

The algorithms are ordered according to the computed ranks for each task.

## 2.1 Dot- and boxplot

*Dot- and boxplots* for visualizing raw assessment data separately for each algorithm. Boxplots representing descriptive statistics over all cases (median, quartiles and outliers) are combined with horizontally jittered dots representing individual cases.

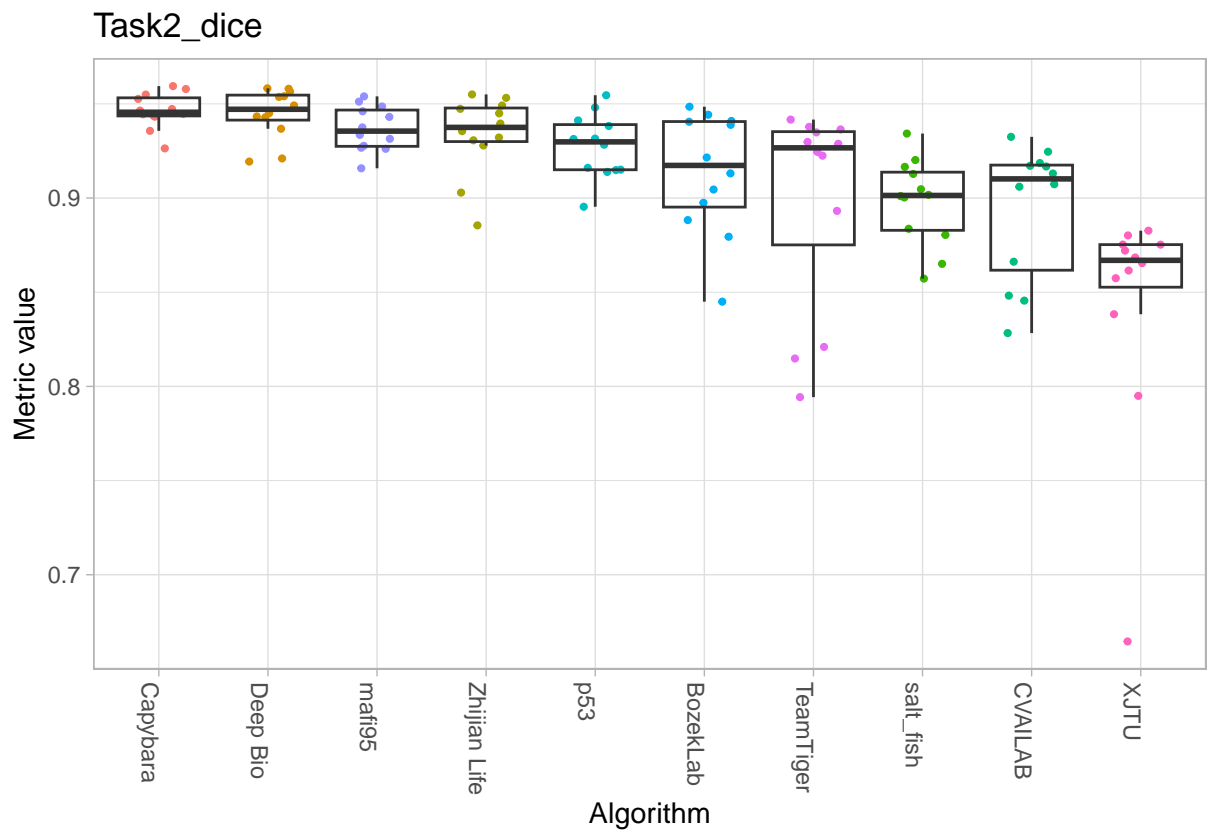

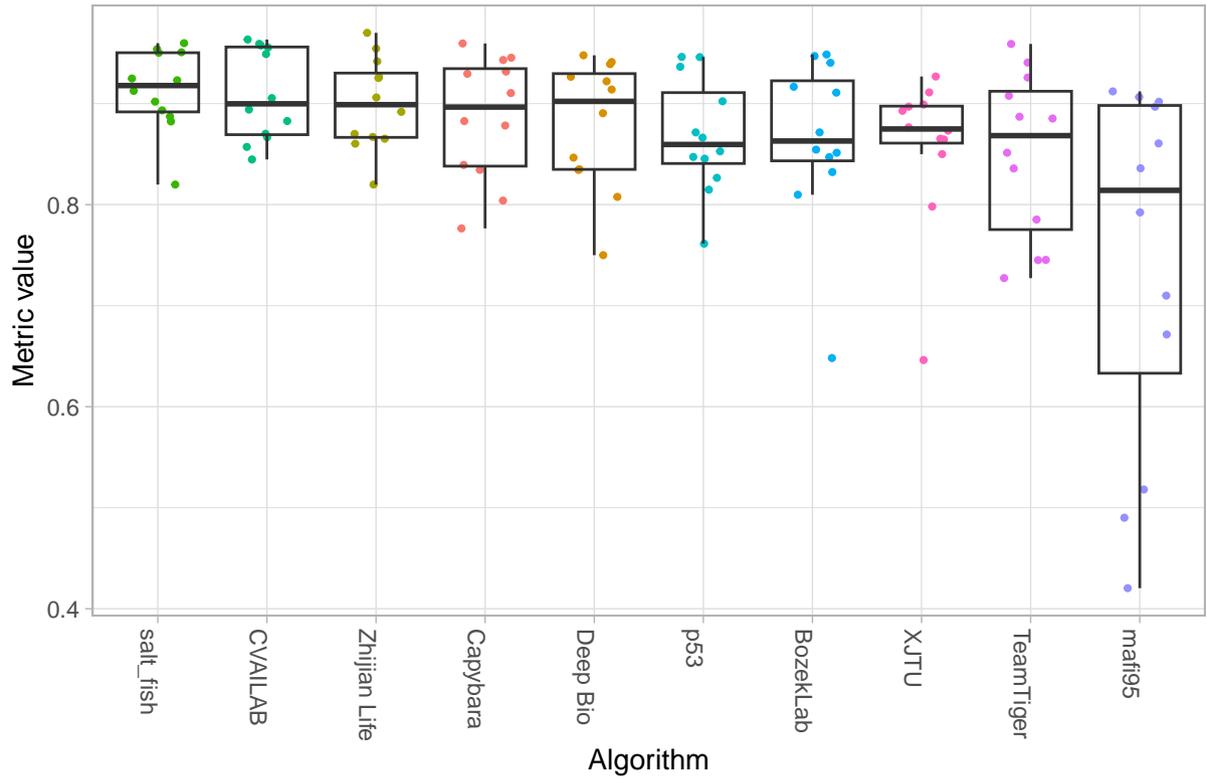

Task2_F1

## 2.2 Podium plot

*Podium plots* (see also Eugster et al., 2008) for visualizing raw assessment data. Upper part (spaghetti plot): Participating algorithms are color-coded, and each colored dot in the plot represents a metric value achieved with the respective algorithm. The actual metric value is encoded by the y-axis. Each podium (here: $p$=10) represents one possible rank, ordered from best (1) to last (here: 10). The assignment of metric values (i.e. colored dots) to one of the podiums is based on the rank that the respective algorithm achieved on the corresponding case. Note that the plot part above each podium place is further subdivided into $p$ "columns", where each column represents one participating algorithm (here: $p = 10$). Dots corresponding to identical cases are connected by a line, leading to the shown spaghetti structure. Lower part: Bar charts represent the relative frequency for each algorithm to achieve the rank encoded by the podium place.

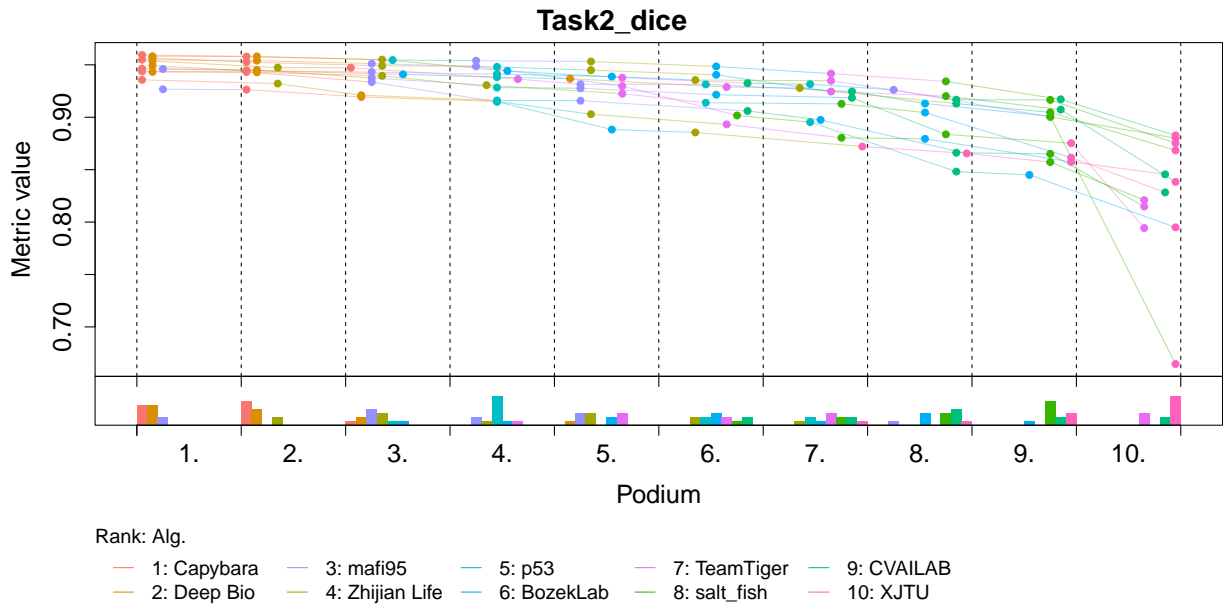

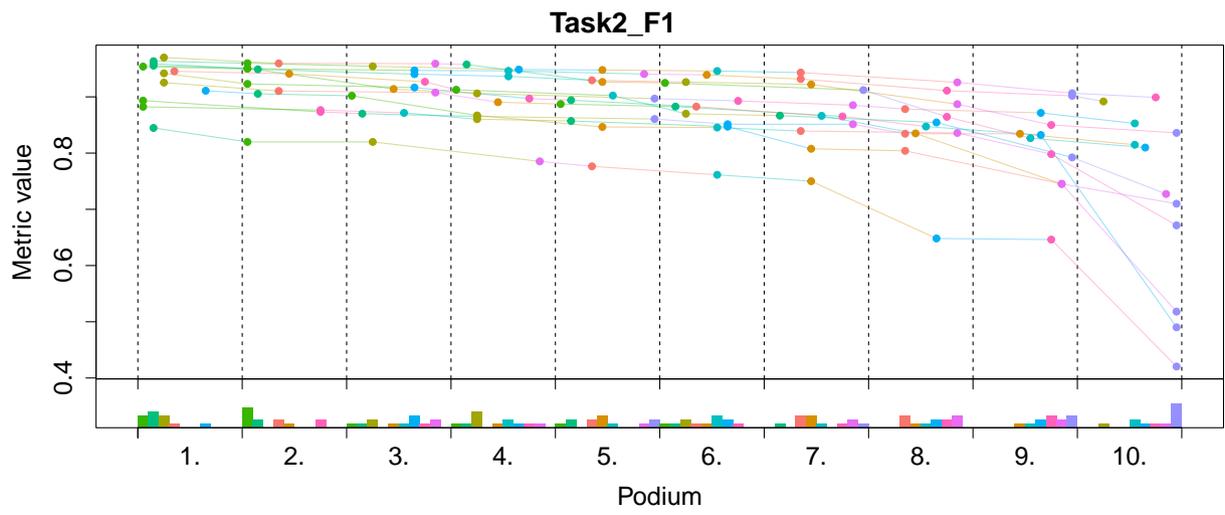

**Task2_F1**

## 2.3 Ranking heatmap

*Ranking heatmaps* for visualizing raw assessment data. Each cell $(i, A_j)$ shows the absolute frequency of cases in which algorithm $A_j$ achieved rank $i$.

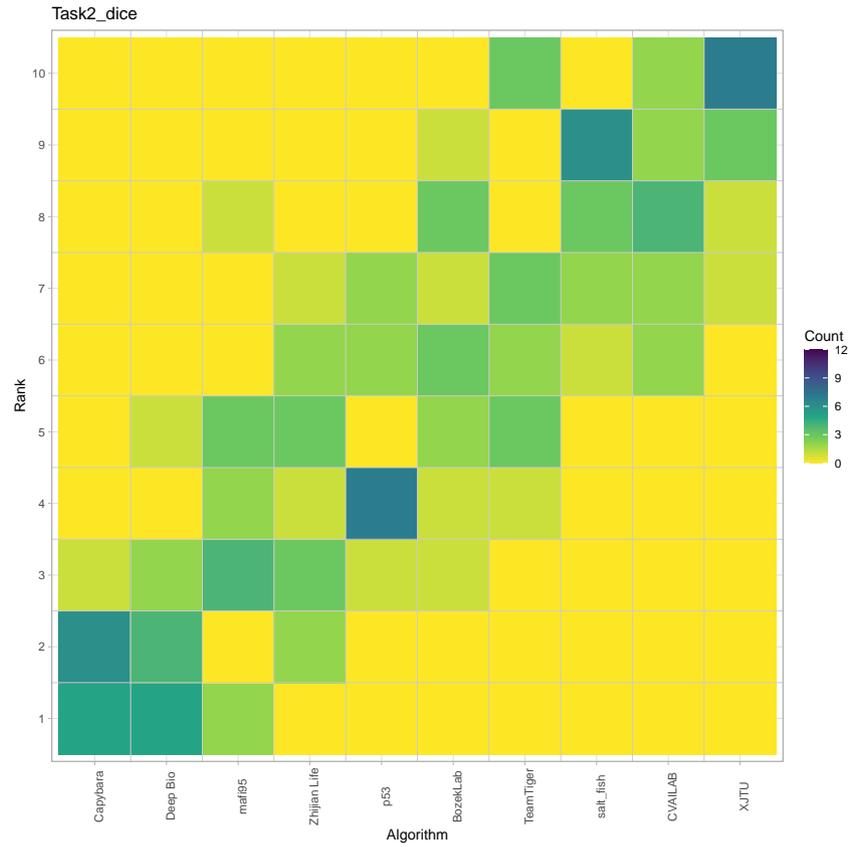

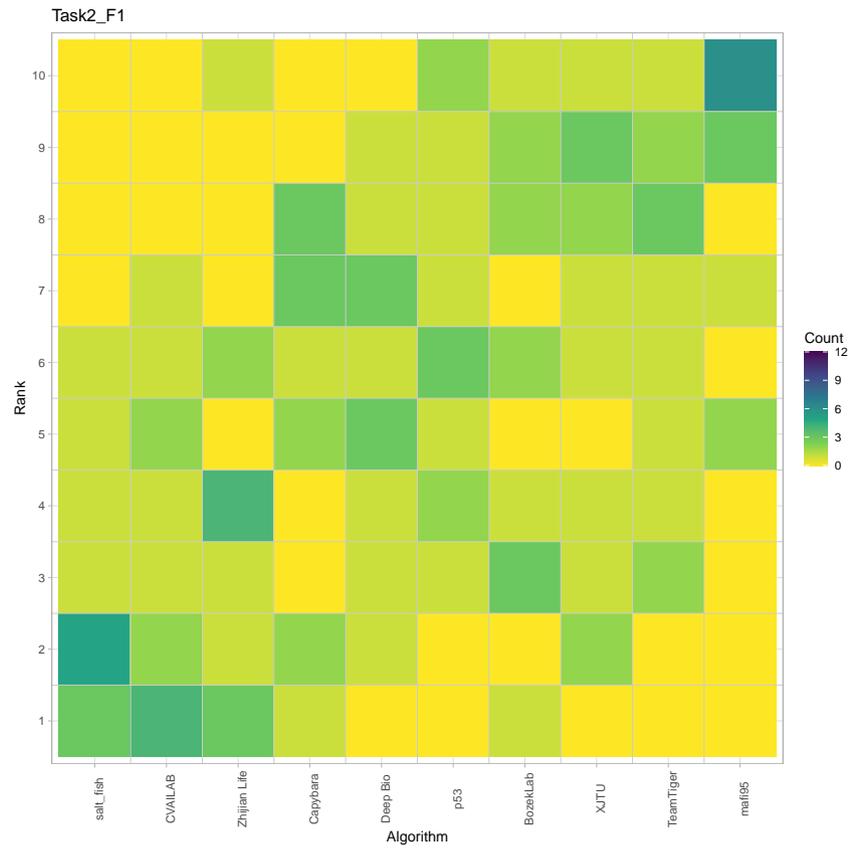

# 3 Visualization of ranking stability

## 3.1 *Blob plot* for visualizing ranking stability based on bootstrap sampling

Algorithms are color-coded, and the area of each blob at position $(A_i, \text{rank } j)$ is proportional to the relative frequency $A_i$ achieved rank $j$ across $b = 1000$ bootstrap samples. The median rank for each algorithm is indicated by a black cross. 95% bootstrap intervals across bootstrap samples are indicated by black lines.

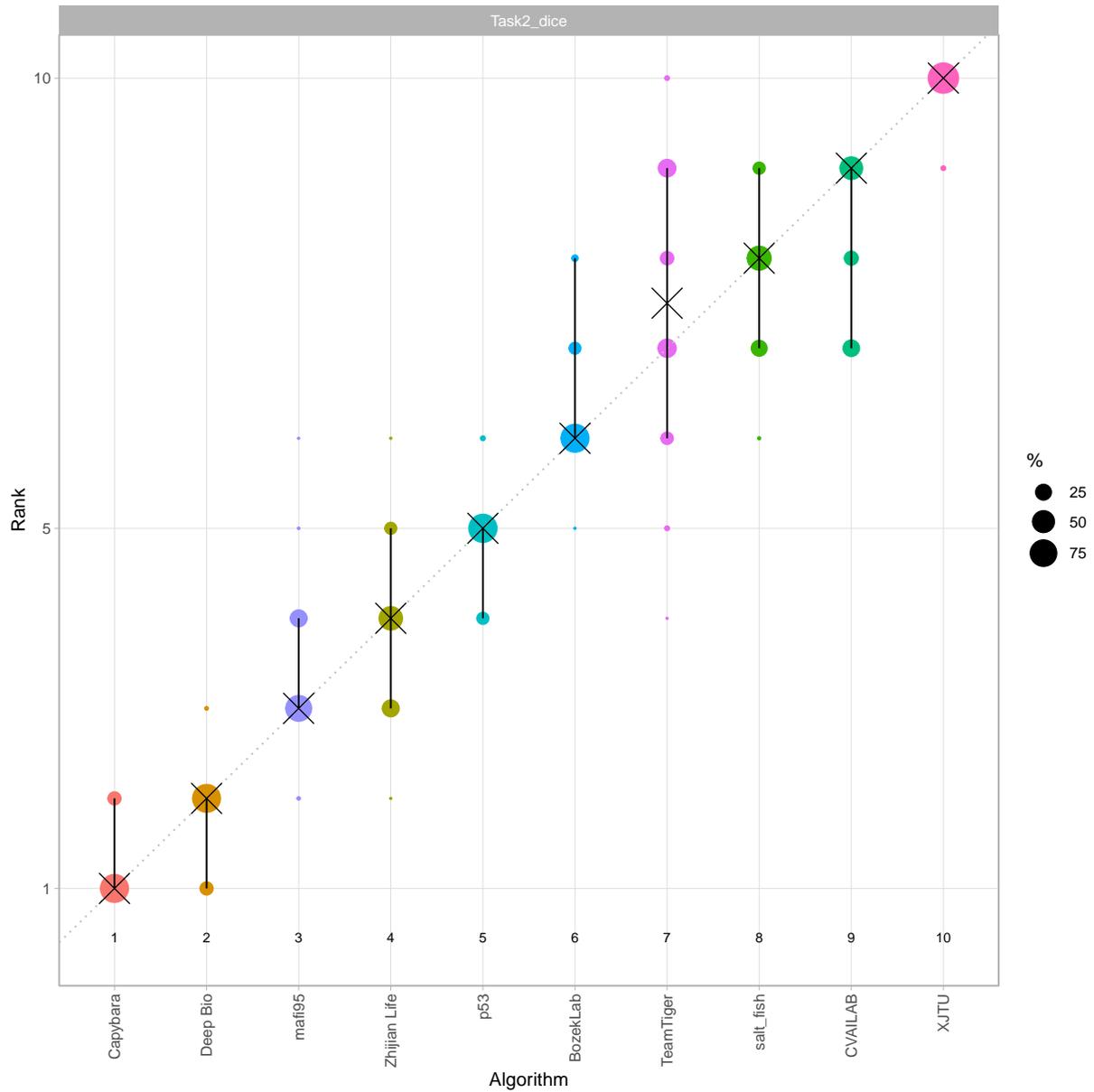

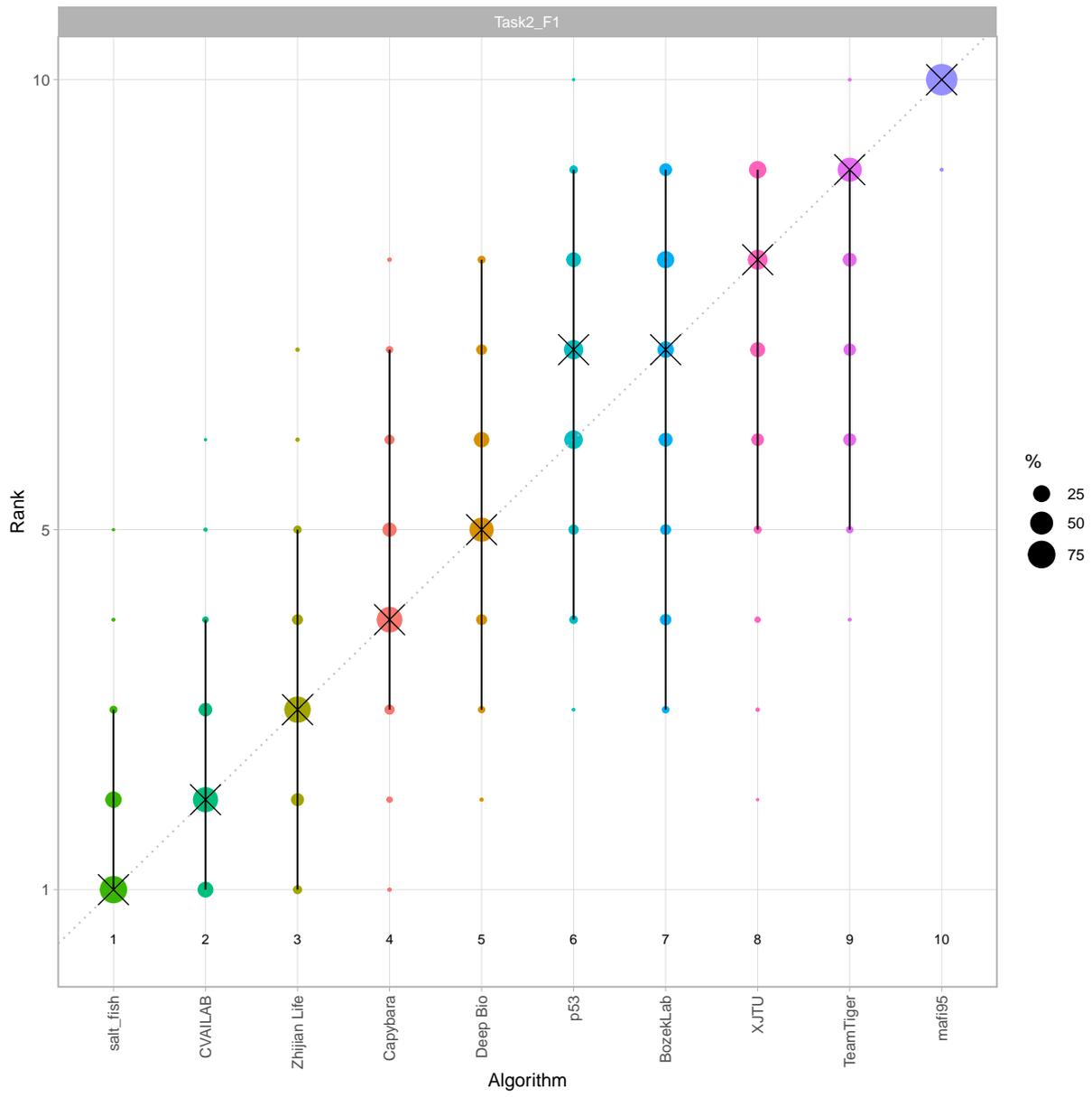

## 3.2 *Violin plot* for visualizing ranking stability based on bootstrapping

The ranking list based on the full assessment data is pairwise compared with the ranking lists based on the individual bootstrap samples (here $b = 1000$ samples). For each pair of rankings, Kendall's $\tau$ correlation is computed. Kendall's $\tau$ is a scaled index determining the correlation between the lists. It is computed by evaluating the number of pairwise concordances and discordances between ranking lists and produces values between $-1$ (for inverted order) and 1 (for identical order). A violin plot, which simultaneously depicts a boxplot and a density plot, is generated from the results.

Summary Kendall's tau:

| Task | mean | median | q25 | q75 |
|------|------|--------|-----|-----|
| Task2_dice | 0.9102222 | 0.9111111 | 0.8666667 | 0.9555556 |
| Task2_F1 | 0.8371111 | 0.8222222 | 0.7777778 | 0.9111111 |

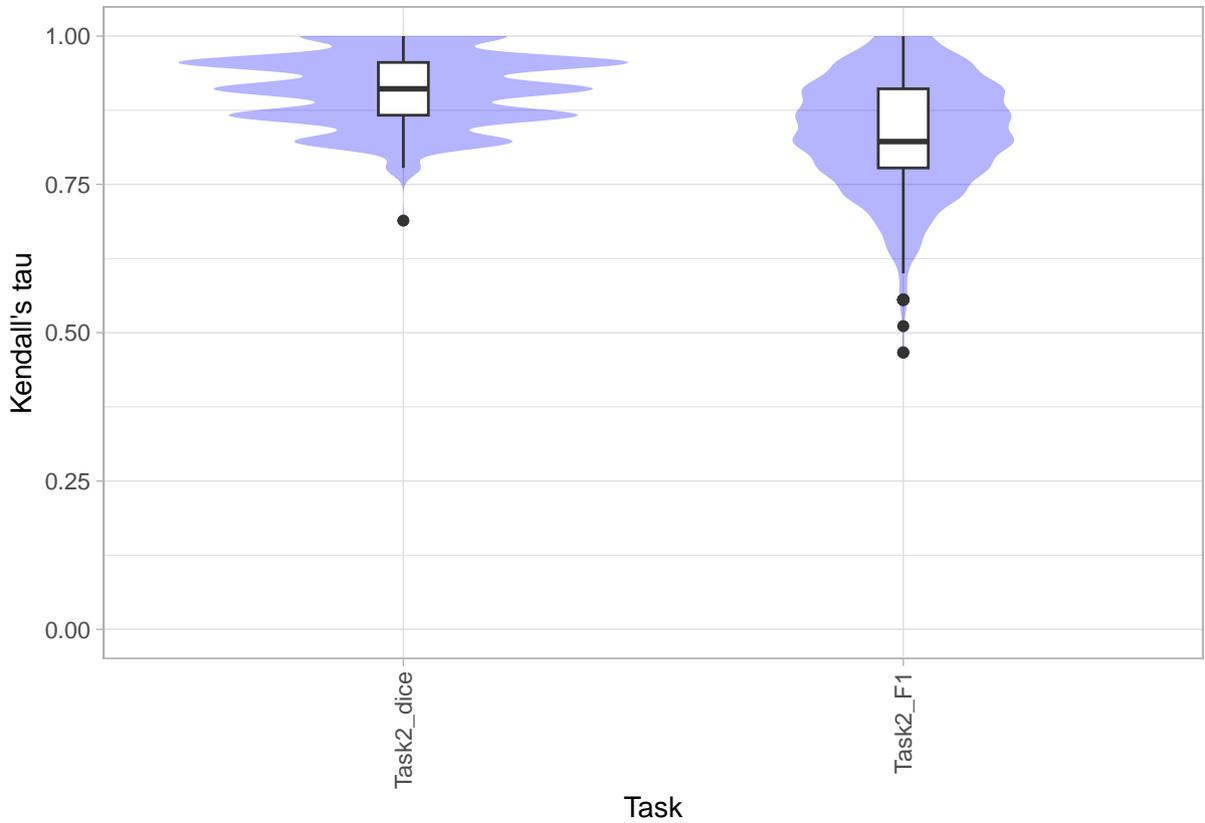

## 3.3 *Significance maps* for visualizing ranking stability based on statistical significance

*Significance maps* depict incidence matrices of pairwise significant test results for the one-sided Wilcoxon signed rank test at a 5% significance level with adjustment for multiple testing according to Holm. Yellow shading indicates that metric values from the algorithm on the x-axis were significantly superior to those from the algorithm on the y-axis, blue color indicates no significant difference.

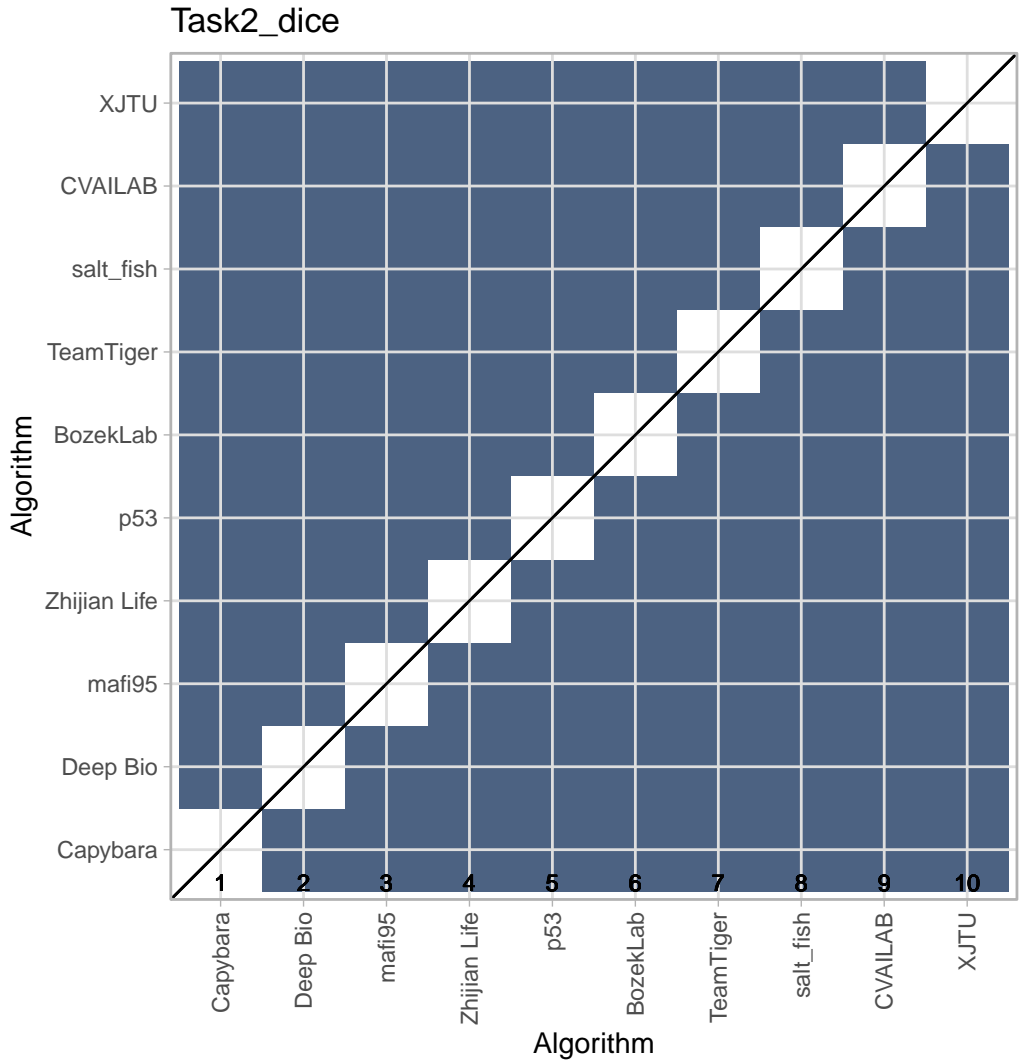

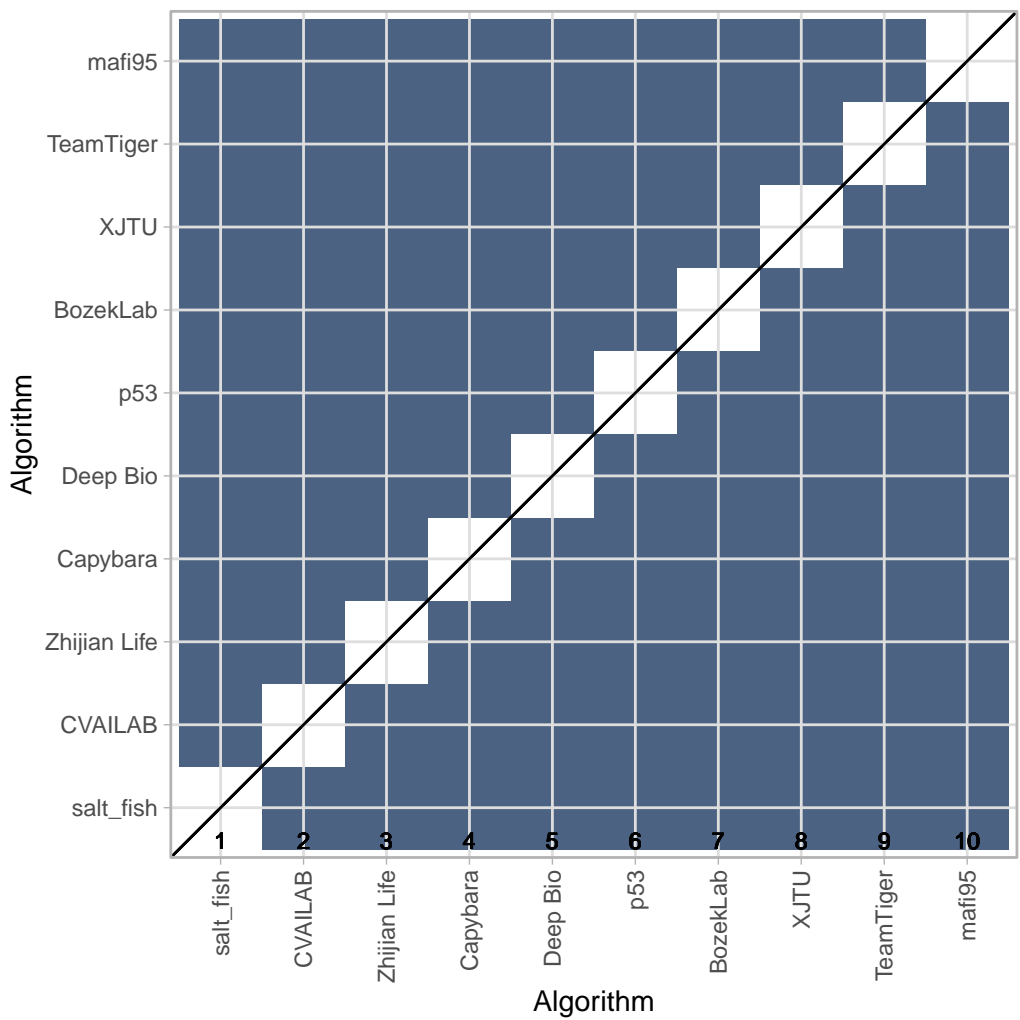

Task2_F1

## 3.4 Ranking robustness to ranking methods

*Line plots* for visualizing ranking robustness across different ranking methods. Each algorithm is represented by one colored line. For each ranking method encoded on the x-axis, the height of the line represents the corresponding rank. Horizontal lines indicate identical ranks for all methods.

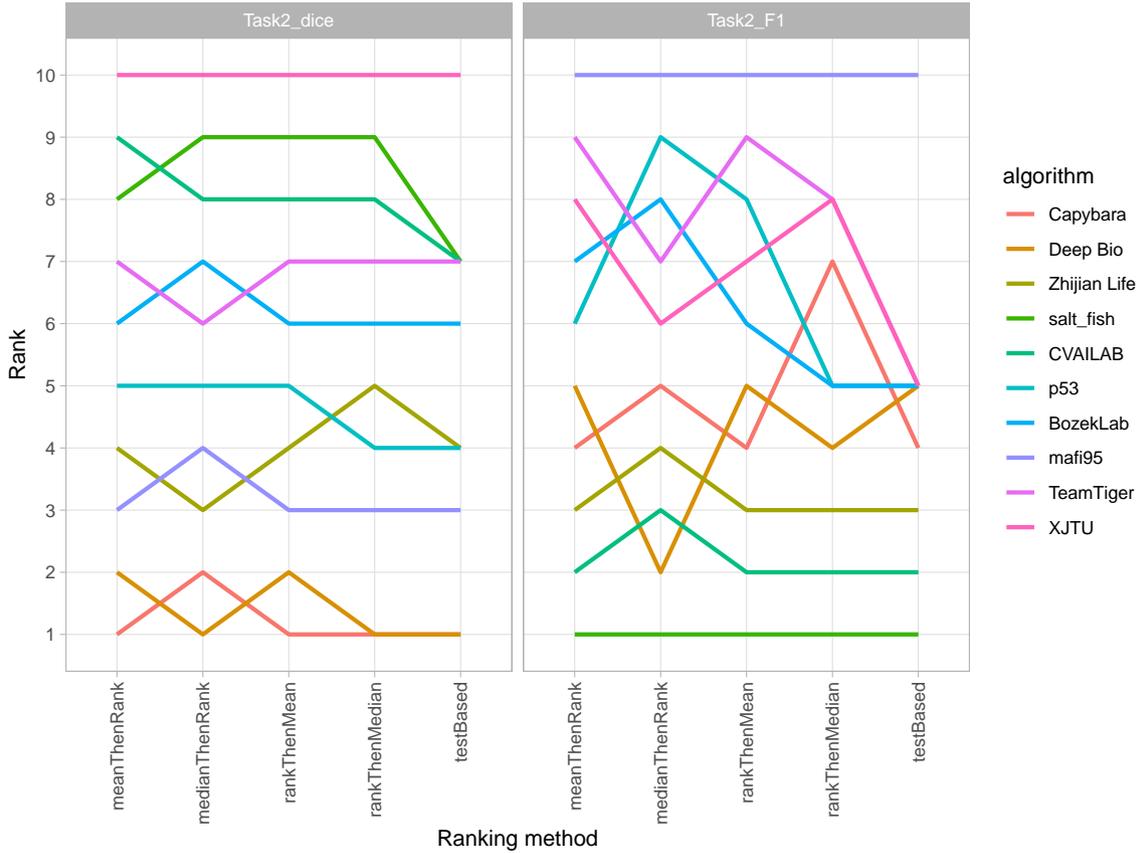

# 4 Visualization of cross-task insights

The algorithms are ordered according to consensus ranking.

## 4.1 Characterization of algorithms

### 4.1.1 Ranking stability: Variability of achieved rankings across tasks

Algorithms are color-coded, and the area of each blob at position $(A_i, \text{rank } j)$ is proportional to the relative frequency $A_i$ achieved rank $j$ across multiple tasks. The median rank for each algorithm is indicated by a black cross. This way, the distribution of ranks across tasks can be intuitively visualized.

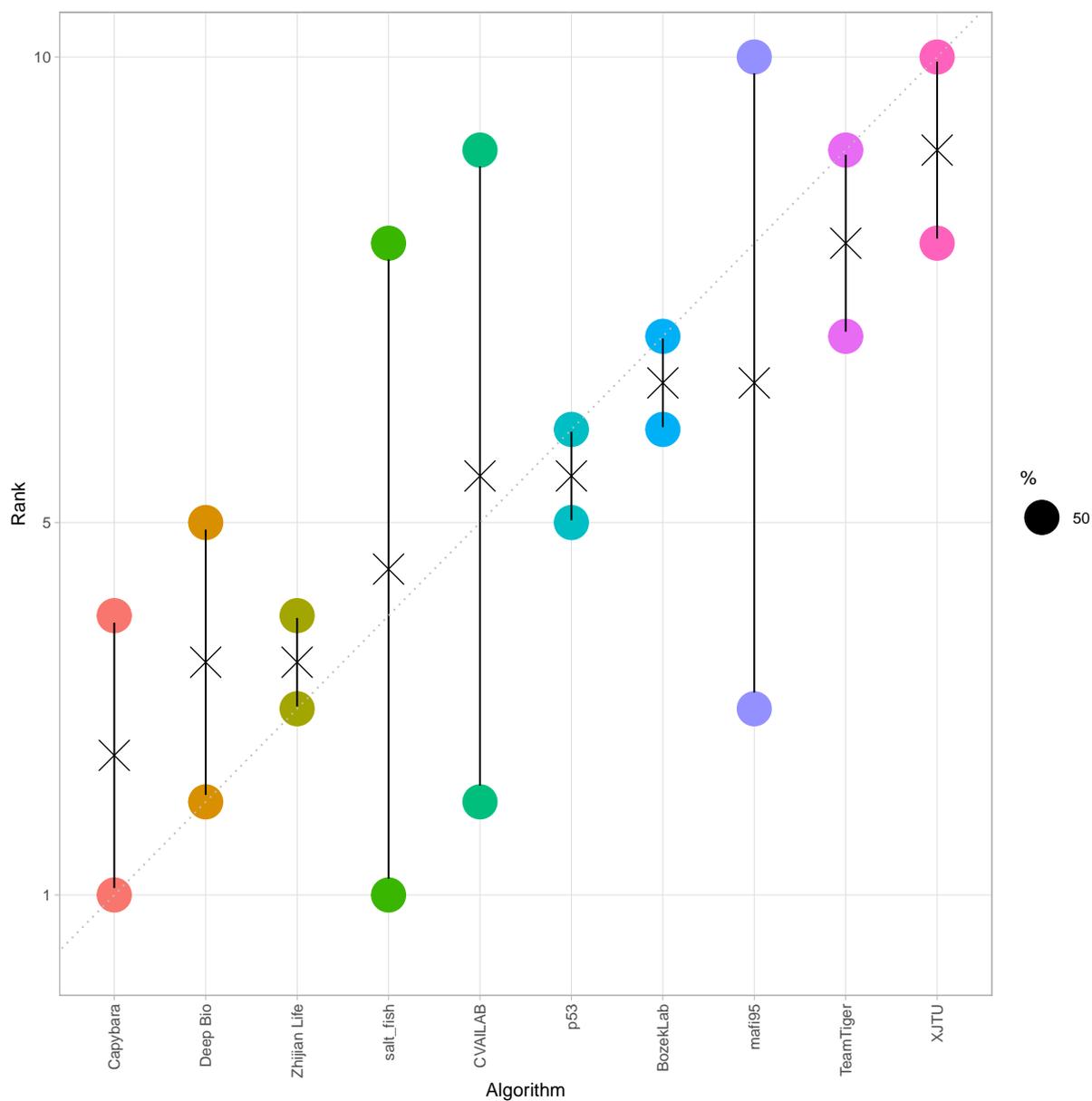

## 4.1.2   Ranking stability: Ranking variability via bootstrap approach

A blob plot of bootstrap results over the different tasks separated by algorithm allows another perspective on the assessment data. This gives deeper insights into the characteristics of tasks and the ranking uncertainty of the algorithms in each task.

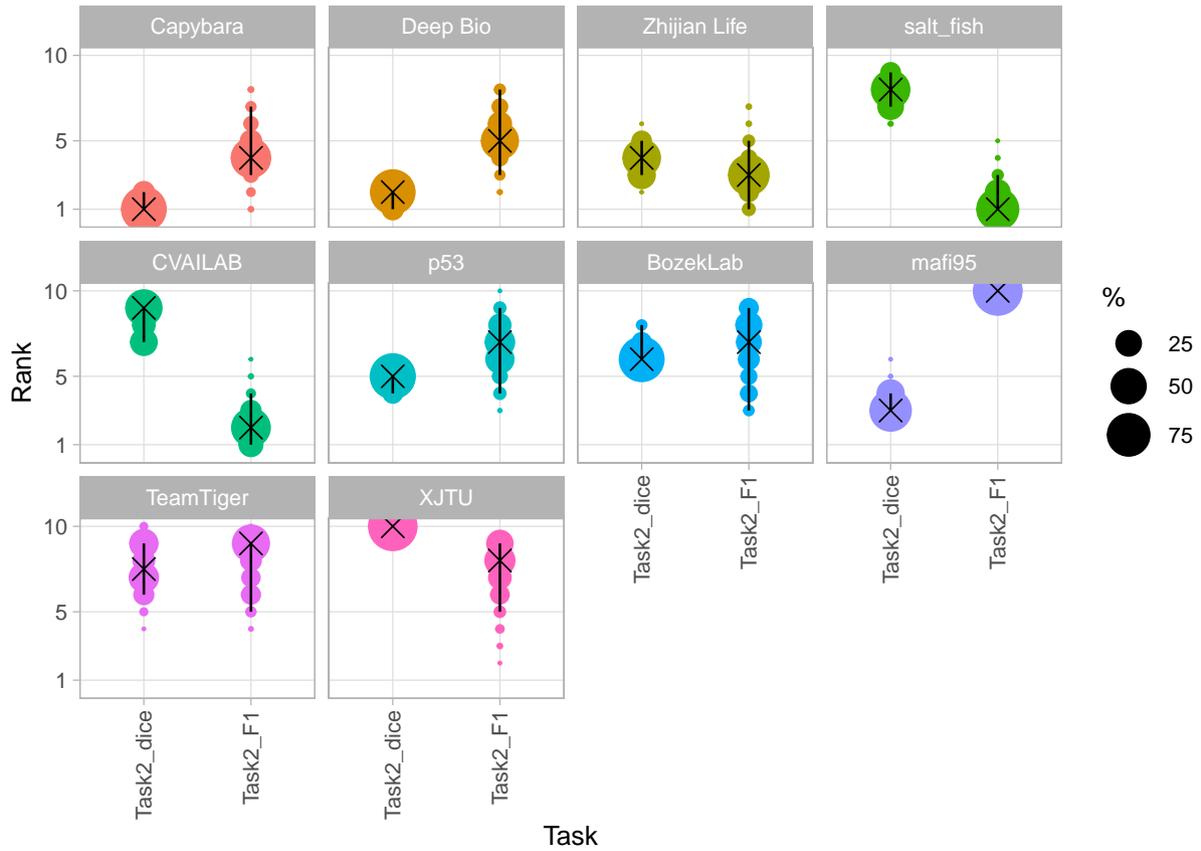

An alternative representation is provided by a stacked frequency plot of the observed ranks, separated by algorithm. Observed ranks across bootstrap samples are displayed with coloring according to the task. For algorithms that achieve the same rank in different tasks for the full assessment data set, vertical lines are on top of each other. Vertical lines allow to compare the achieved rank of each algorithm over different tasks.

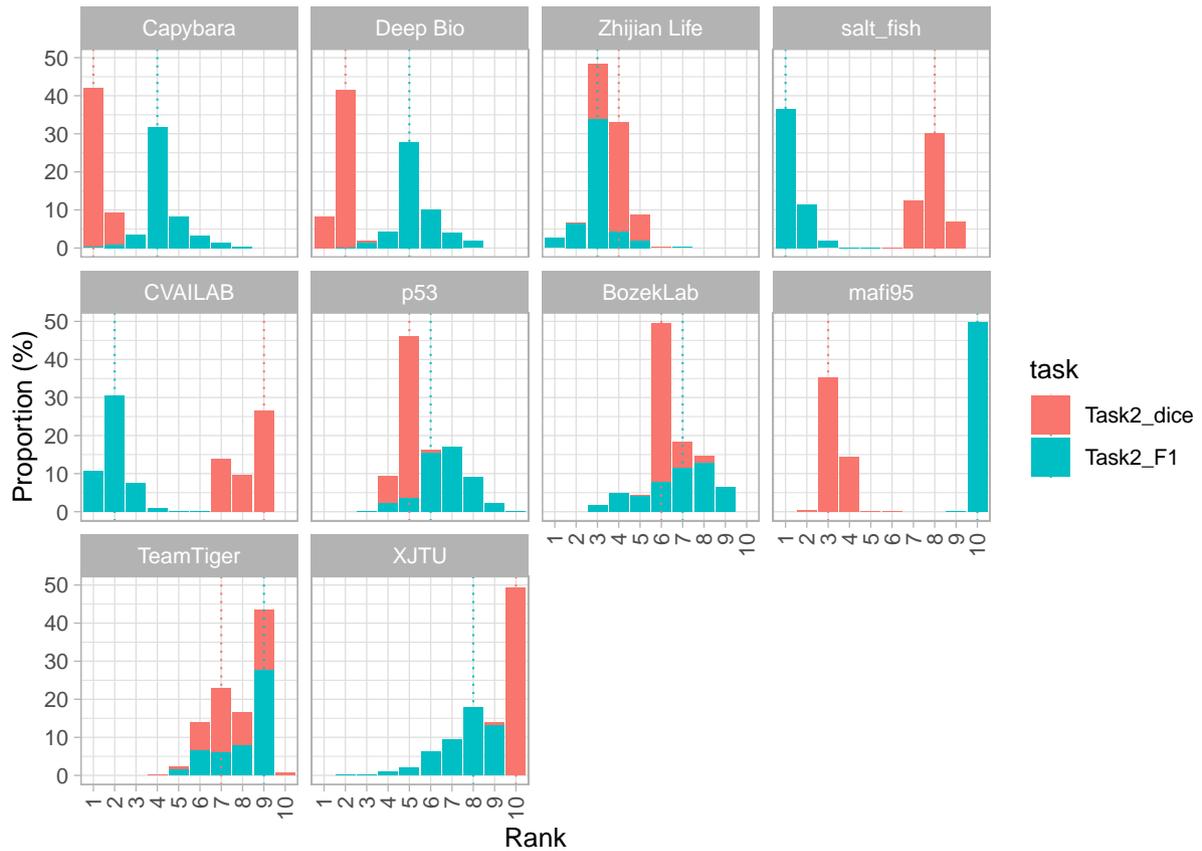

## 4.2 Characterization of tasks

### 4.2.1 Visualizing bootstrap results

To investigate which tasks separate algorithms well (i.e., lead to a stable ranking), a blob plot is recommended.

Bootstrap results can be shown in a blob plot showing one plot for each task. In this view, the spread of the blobs for each algorithm can be compared across tasks. Deviations from the diagonal indicate deviations from the consensus ranking (over tasks). Specifically, if rank distribution of an algorithm is consistently below the diagonal, the algorithm performed better in this task than on average across tasks, while if the rank distribution of an algorithm is consistently above the diagonal, the algorithm performed worse in this task than on average across tasks. At the bottom of each panel, ranks for each algorithm in the tasks are provided.

Same as in Section 3.1 but now ordered according to consensus.

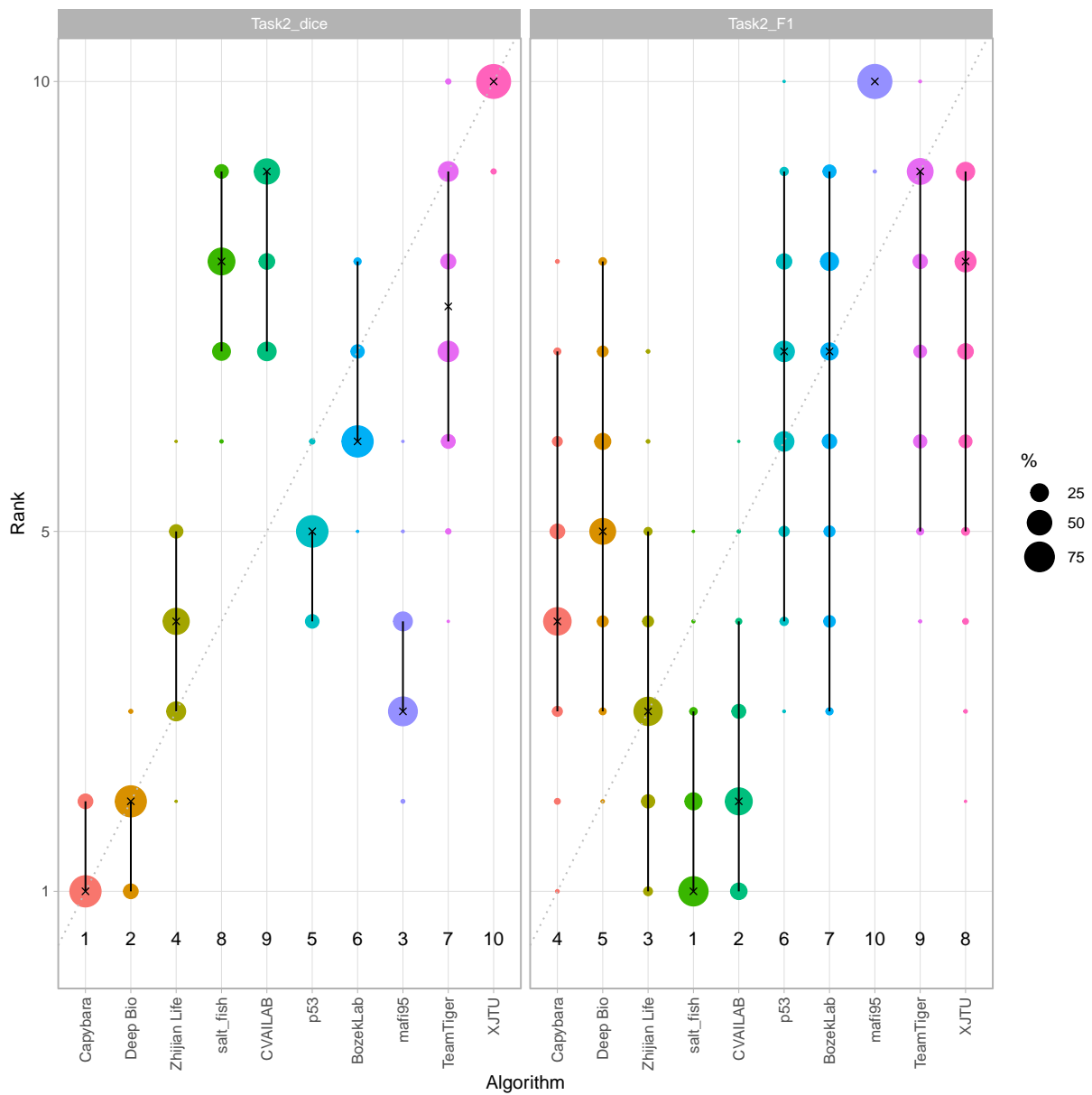

### 4.2.2 Cluster Analysis

Dendrogram from hierarchical cluster analysis and *network-type graphs* for assessing the similarity of tasks based on challenge rankings.

A dendrogram is a visualization approach based on hierarchical clustering. It depicts clusters according to a chosen distance measure (here: Spearman's footrule) as well as a chosen agglomeration method (here: complete and average agglomeration).

```
##
## Cluster analysis only sensible if there are >2 tasks.
```

# 5  References

Wiesenfarth, M., Reinke, A., Landman, B.A., Eisenmann, M., Aguilera Saiz, L., Cardoso, M.J., Maier-Hein, L. and Kopp-Schneider, A. Methods and open-source toolkit for analyzing and visualizing challenge results. *Sci Rep* **11**, 2369 (2021). https://doi.org/10.1038/s41598-021-82017-6

M. J. A. Eugster, T. Hothorn, and F. Leisch, "Exploratory and inferential analysis of benchmark experiments," Institut fuer Statistik, Ludwig-Maximilians-Universitaet Muenchen, Germany, Technical Report 30, 2008. [Online]. Available: http://epub.ub.uni-muenchen.de/4134/.

 

\newpage
 \renewcommand{\thesection}{\Alph{section}}
\end{document}